\documentclass[10pt,twocolumn,letterpaper]{article}

\usepackage[pagenumbers]{cvpr} 

\usepackage{kotex}
\usepackage{times}
\usepackage{epsfig}
\usepackage{graphicx}
\usepackage{amsmath}
\usepackage{amssymb}
\usepackage{booktabs}
\usepackage{kotex}
\usepackage{multirow}
\usepackage{makecell}
\usepackage{color, colortbl}
\usepackage{caption}
\usepackage{soul}
\usepackage{booktabs}
\usepackage[normalem]{ulem}
\usepackage{setspace}
\usepackage[accsupp]{axessibility}

\def\MethodName{CLIPtone}
\def\OurHypernet{text adapter}

\newcommand{\Eq}[1]   {Eq.\ (#1)}

\newcommand{\Fig}[1]  {Fig.\ #1}

\newcommand{\Tbl}[1]  {Tab.\ #1}

\renewcommand{\paragraph}[1]{\vspace{2pt}\noindent\textbf{#1}~~}
\usepackage[dvipsnames]{xcolor}
\definecolor{cvprblue}{rgb}{0.21,0.49,0.74}
\usepackage[pagebackref,breaklinks,colorlinks,citecolor=cvprblue]{hyperref}

\title{\MethodName{}: Unsupervised Learning for 
Text-based Image Tone Adjustment}

\author{Hyeongmin Lee$^{1,\ast}$ ~ ~ ~
Kyoungkook Kang$^{2,\ast}$ ~ ~ ~
Jungseul Ok$^{1,2}$ ~ ~ ~
Sunghyun Cho$^{1,2}$\\[2mm]
POSTECH $^{1}$GSAI \& $^{2}$CSE \\
{\tt\small \{hmin970922, kkang831, jungseul, s.cho\}@postech.ac.kr}
}

\newcommand\blfootnote[1]{%
  \begingroup
  \renewcommand\thefootnote{}\footnote{#1}%
  \addtocounter{footnote}{-1}%
  \endgroup
}

\spacing{0.93}
\begin{document}

\twocolumn[{
\renewcommand\twocolumn[1][]{#1}
\maketitle
\begin{center}
    \centering
    \vspace{-0.5cm}
    \captionsetup{type=figure}
    \scalebox{0.95}{\includegraphics[width=\textwidth]{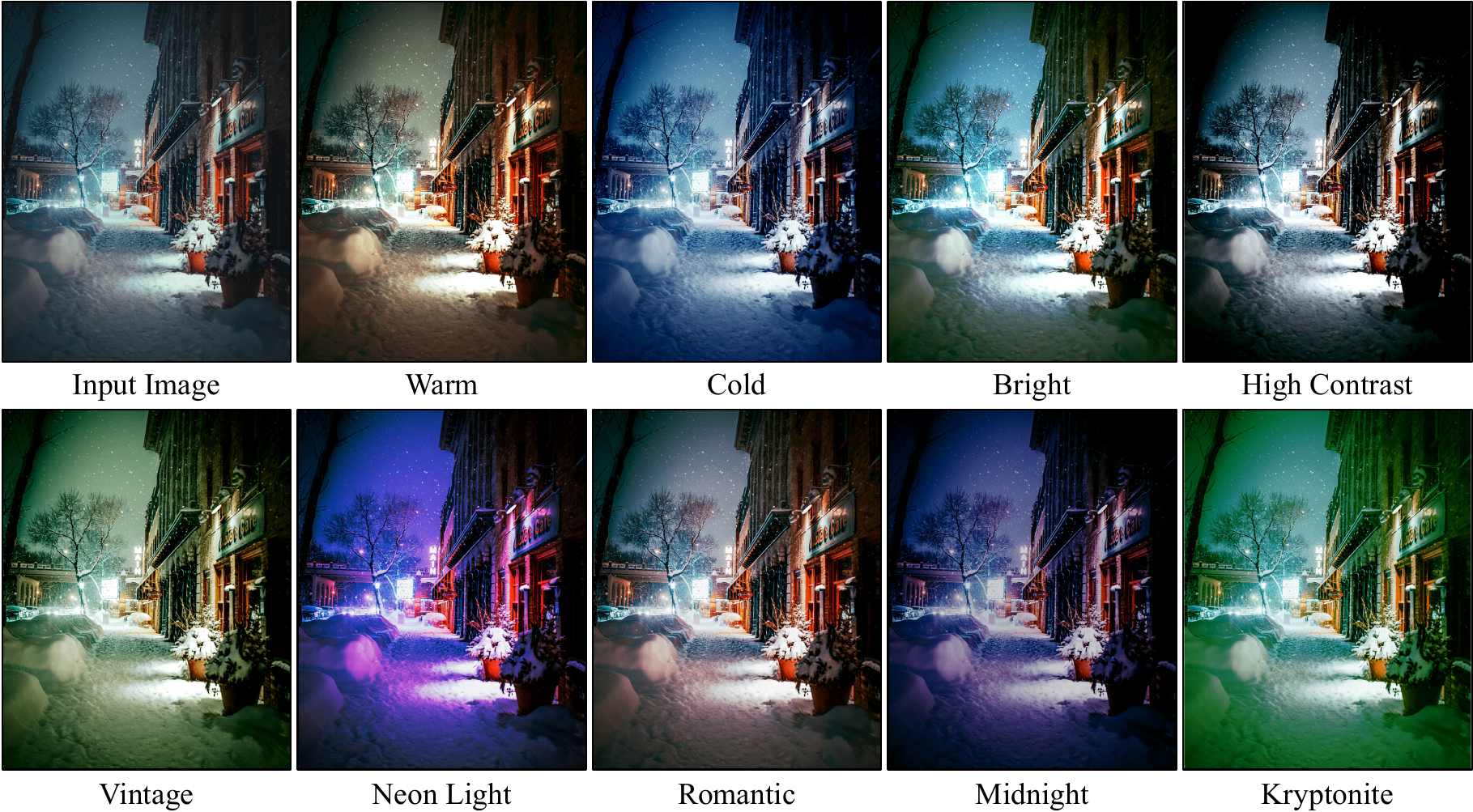}}
    \vspace{-0.3cm}
    \captionof{figure}{
    We present \MethodName{}, a text-based image tone adjustment framework trained in an unsupervised manner.
    With its superior understanding of natural languages, \MethodName{} is capable of performing successful adjustments across a range of text descriptions, including those previously deemed challenging.
    }
    \label{fig:fig1_teaser}
    \vspace{-0.04cm}
\end{center}
}]

\begin{abstract}
\vspace{-0.3cm}
Recent image tone adjustment (or enhancement) approaches have predominantly adopted supervised learning for learning human-centric perceptual assessment. However, these approaches are constrained by intrinsic challenges of supervised learning.
Primarily, the requirement for expertly-curated or retouched images escalates the data acquisition expenses. 
Moreover, their coverage of target styles is confined to stylistic variants inferred from the training data. 
To surmount the above challenges, we propose an unsupervised learning-based approach for text-based image tone adjustment, \MethodName{}, that extends an existing image enhancement method to accommodate natural language descriptions.
Specifically, we design a hyper-network to adaptively modulate the pretrained parameters of a backbone model based on a text description. 
To assess whether an adjusted image aligns with its text description without a ground-truth image, we utilize CLIP, which is trained on a vast set of language-image pairs and thus encompasses the knowledge of human perception.
The major advantages of our approach are threefold: (\lowercase\expandafter{\romannumeral1}) minimal data collection expenses, (\lowercase\expandafter{\romannumeral2}) support for a range of adjustments, and (\lowercase\expandafter{\romannumeral3}) the ability to handle novel text descriptions unseen in training.
The efficacy of the proposed method
is demonstrated through comprehensive experiments including a user study.
\vspace{-0.8cm}

\end{abstract}
\blfootnote{$^{\ast}$ Equal contribution}    
\section{Introduction}
\label{sec:intro}

Image tone adjustment aims for the alteration of the tonal properties of an image, including brightness, contrast, and color balance. 
It is also termed image tone enhancement, as it has primarily been exploited to enhance the image aesthetics.
It is essential for various applications ranging from photography to medical imaging, and the importance of effective tone adjustment techniques has grown significantly.

Recent learning-based approaches~\cite{gharbi2017deep, moran2020deeplpf, he2020conditional, wang2021real} predominantly rely on pairwise datasets and supervised learning to learn perceptual adjustments aligned with human perception.
Representative datasets like MIT-Adobe 5K~\cite{ma5k} and PPR10K~\cite{ppr10k}, which comprise pairs of an image and its expert-retouched version, have emerged, and several methods~\cite{he2020conditional, kim2021representative, moran2021curl, yang2022adaint} have been proposed to learn the mappings between such pairs of images.
However, such supervised learning-based methodologies bear several limitations.
First, they demand cost-intensive datasets stemming from expert retouching.
Second, they are confined to limited amounts of adjustments that are inferred from the datasets.
Lastly, they allow for only automatic image enhancement and lack the capability for diverse adjustments with any guidance, as the datasets are composed of paired images without specific guidance annotations.

To reduce the burden of collecting paired images, a few studies~\cite{hu2018exposure, chen2018deep, ignatov2018wespe, kim2020global, zeng2020learning} suggest weakly-supervised learning-based approaches to learn the mapping from the unpaired images.
They predominantly leverage the adversarial learning of GANs~\cite{gan} to capture the tonal properties of a target image set. 
Although they do use unpaired data, there is still a requirement to manually define distinct sets for both source and target images.
In addition, they also share the other limitations of the supervised learning-based approaches.

Meanwhile, a handful of works~\cite{t2onet, cagan} introduce text-based approaches to enable controllable adjustments utilizing text descriptions.
To this end, they construct datasets consisting of triplets (i.e., a source image, a target image, and a text description on the adjustment) and train neural networks on them.
However, due to the reliance on a limited range of descriptions in the datasets, they cannot accommodate natural language descriptions beyond them.

In this paper, we propose a novel text-based image tone adjustment framework, \MethodName{}, that is trained in an unsupervised manner requiring only source images and any tone-related text descriptions without the need for paired images.
Our core insight is that the assessment of perceptual adjustments can be achieved with the recent language-image representation model, CLIP~\cite{clip}, as it is trained on a vast amount of language-image pairs and thus encapsulates the human perception.
Grounded on this insight, we extend an existing image enhancement model to enable varied adjustments based on text descriptions.
Specifically, we design a hyper-network to adaptively modulate the pretrained parameters of a backbone model based on a text description.
In order for our network to learn tone adjustment in an unsupervised manner, we further introduce a CLIP directional loss and regularization losses.

\MethodName{} enjoys several unique benefits
stemming from introducing CLIP as criterion of human perception.
It necessitates only arbitrary images and tone-related text descriptions for its training, which can be collected with minimal costs.
It also supports vast amounts of adjustments previously deemed challenging with text descriptions, as shown in \Fig{\ref{fig:fig1_teaser}}, thanks to CLIP's comprehensive understanding of natural language.
Lastly, it is capable of handling novel text descriptions unseen in training.

Recently, in line with remarkable capabilities of contemporary generative models, numerous studies~\cite{meng2021sdedit, gal2022stylegan,crowson2022vqgan, brooks2023instructpix2pix} have been proposed dictating the attributes of the generated images with text descriptions.
Among them, our work shares similarities with text-based image colorization and text-based image editing approaches, in the aspect of modifying input images based on text descriptions.
Compared to these methods, \MethodName{} stands out for its remarkably fewer parameters and shorter inference time. 

Our contributions are summarized as follows:
\begin{itemize}
    \item We present, for the first time, an unsupervised image tone adjustment framework \MethodName{}, that captures human perception on tonal properties utilizing CLIP, enabling a range of adjustments and zero-shot prediction.
    \item We design a hyper-network to effectively modulate the pretrained parameters of an existing image enhancement network and suggest training strategies tailored for unsupervised learning of \MethodName{}.
    \item Through extensive experiments, we demonstrate that our approach is lighter, faster, and more effective compared to existing methods.
\end{itemize}
\section{Related Work}
\label{sec:relatedwork}
\paragraph{Image Tone Adjustment Methods.}
To improve the aesthetic quality of images, a range of learning-based automatic tone adjustment (or enhancement) methods have been developed.
They have explored optimal neural architectures to directly estimate dense pixel values~\cite{ignatov2017dslr, moran2020deeplpf, moran2021curl} or the parameters of traditional physical models such as color transform matrices~\cite{yan2016automatic, gharbi2017deep, chai2020supervised}, and Look-Up Tables (LUTs)~\cite{he2020conditional, kim2021representative, zeng2020learning, wang2021real, yang2022adaint, yang2022seplut, zhang2022clut, liu20234d}.  
Among these, methods exploiting 3D LUTs~\cite{zeng2020learning, wang2021real, yang2022adaint, yang2022seplut, zhang2022clut, liu20234d} have garnered attention for their superior capacity and efficiency.
Since Zeng~\etal~\cite{zeng2020learning} have proposed adaptive 3D LUTs to input images, Wang~\etal~\cite{wang2021real} proposed spatially-aware 3D LUTs for local adjustment, and Yang~\etal~\cite{yang2022adaint} increased the capacities of 3D LUTs through non-uniform sampling.

For neural networks to learn adjustments aligned with human perception, supervised learning-based approaches~\cite{gharbi2017deep, moran2020deeplpf, he2020conditional, wang2021real} have predominantly been attempted. They rely on paired datasets such as MIT-Adobe 5K \cite{ma5k} and PPR10K \cite{ppr10k}, which consist of source images and their expert-retouched counterparts.
To alleviate the collection costs associated with expert retouching, weakly-supervised learning-based approaches ~\cite{chen2018deep, hu2018exposure, kim2020global, zeng2020learning} have been introduced. 
They use unpaired data comprised of separate source and target sets of images. These methods primarily leverage the adversarial loss from GANs~\cite{gan} to capture the tonal properties of target sets, and the cycle-consistency loss~\cite{cyclegan} for preserving the structures of source images.

A few methods~\cite{t2onet, cagan} employ text descriptions as conditional inputs to enable diverse and controllable adjustment beyond automatic adjustment.
They construct training datasets consisting of triplets: a source image, a target image, and a text description on the adjustment and train a network on these datasets.
Specifically, Shi~\etal~\cite{t2onet} annotated requests to each pair of the MIT-Adobe 5K dataset~\cite{ma5k}, and Jiang~\etal~\cite{cagan} further proposed dataset augmentation strategies for imbalanced annotations.  

Despite the aforementioned efforts, existing tone adjustment methods have limitations including high data collection costs (even for curating target sets), confined adjustments to stylistic variations of training datasets.
In contrast, \MethodName{} is independent of these issues thanks to leveraging CLIP as a perceptual criterion.

\paragraph{Text-driven Generative Models.}
Recently, cutting-edge generative models have found their uses in image-to-image translation tasks such as image colorization~\cite{huang2022unicolor, l-cad} and editing~\cite{avrahami2022blended,crowson2022vqgan,brooks2023instructpix2pix}.
Early approaches use GANs~\cite{gan} trained on specific domains to manipulate real images through GAN inversion techniques~\cite{richardson2021encoding, tov2021designing, bdinvert, alaluf2021restyle}, changing their latent vectors~\cite{xia2021tedigan,bau2021paint, patashnik2021styleclip} or adapting generator networks~\cite{li2020manigan,gal2022stylegan} to different domains according to text descriptions.
More recently, with advances in auto-regressive models~\cite{vqgan} and diffusion models~\cite{diffusion, song2020improved, ddpm}, generative models trained on large amounts of text-image pairs have been showing superior performance in image synthesis~\cite{stablediffusion}. Several studies have proposed to finetune these models to manipulate images based on text descriptions~\cite{controlnet, l-cad, brooks2023instructpix2pix} benefiting from their superior generative prior.
While these image manipulation methods seem to have similarities with our methods, they struggle to maintain the structures of original images and require excessive time for manipulation.
\begin{figure}[t]
  \centering
    \includegraphics[width=\linewidth]{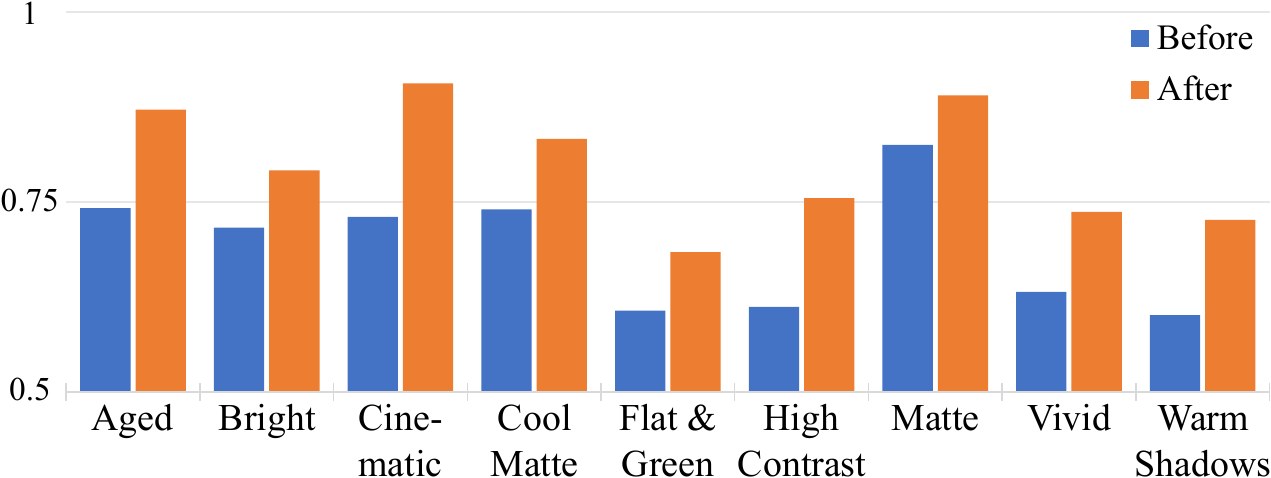}
    \vspace{-0.6cm}
   \caption{
   We apply tone adjustment filters from Adobe Lightroom Classic~\cite{adobelightroomclassic} to 500 images from the MIT-Adobe 5K dataset~\cite{ma5k},
   and calculate the relative similarities between the images and filter names in the CLIP space.
   For all filters, the filtered images have higher similarity scores than the source images, implying that CLIP can assess the tonal properties of images in a manner aligning with human perception.
}   
   \label{fig:fig2_clipgraph}
   \vspace{-0.4cm}
\end{figure}
\begin{figure*}[t!]
    \centerline{\includegraphics[width=0.92\textwidth]{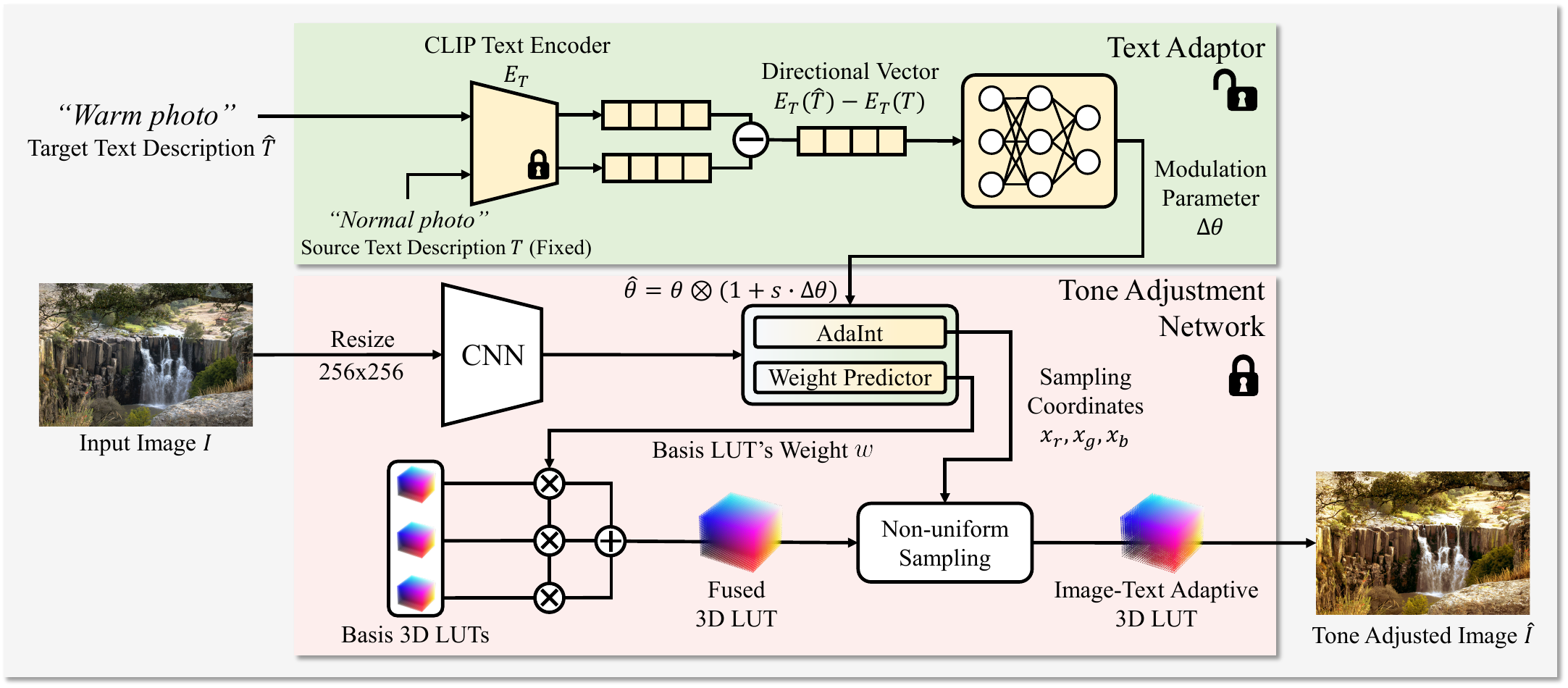}}
    \vspace{-0.3cm}
    \caption{\MethodName{} consists of a \OurHypernet{} and a tone adjustment network.
    From a target text description, the \OurHypernet{} calculates a directional vector within the CLIP embedding space from the source to target text descriptions and estimates the modulation parameter $\Delta\theta$ for the AdaInt module and the weight predictor of the tone adjustment network.
    The modulated tone adjustment network adaptively constructs an image-text adaptive 3D LUT through fusing basis 3D LUTs and non-uniform sampling, ultimately adjusting the color values of an input image.
    }
    \vspace{-0.4cm}
    \label{fig:fig3_framework}
\end{figure*}

\section{CLIP as a Perceptual Criterion}
\label{sec:method}
As mentioned in \cref{sec:intro}, we aim to train a text-based tone adjustment network without paired images by leveraging CLIP~\cite{clip}.
This is based on a hypothesis that CLIP inherently encapsulates information about human perception learned from a vast amount of image-language pairs so that it can be employed for assessing tone adjustment in a manner consistent with human perception.

To verify our hypothesis, we conduct an experiment to examine whether CLIP can correctly assess images whose tones are already known.
To this end, we adopt tone adjustment filters manually designed by experts.
Specifically, we randomly sample 500 images from the MIT-Adobe 5K dataset~\cite{ma5k}, and apply nine filters such as ``aged'', ``bright'', ``cinematic'' from Adobe Lightroom Classic~\cite{adobelightroomclassic},
and construct a dataset consisting of source and filtered image pairs for each filter.
In each dataset, the source images have natural tones while the filtered images have specific tones produced by each filter.
We then compute their similarities to their corresponding filter names in the CLIP embedding space.
Specifically, adopting the idea of CLIP-IQA~\cite{clipiqa}, we compute relative similarities defined as:
\begin{equation}
    S(I,T) = \mathrm{softmax}(\textrm{sim}(I, T), \textrm{sim}(I, T'))
\end{equation}
where $I$ and $T$ are an image and a filter name, respectively. $\textrm{sim}(I,T)$ is the cosine similarity between the CLIP embeddings of $I$ and $T$.
$T'$ is a text description ``normal photo''.
To obtain CLIP embeddings, we use the CLIP-RN50 model in this experiment.

\Fig{\ref{fig:fig2_clipgraph}} illustrates the result where the filtered images have higher similarity scores than the source images for all filters.
This result indicates that CLIP holds the potential to serve as a differentiable perceptual criterion for tone adjustment evaluation.
Additional results employing other CLIP models are included in the supplementary material.

\section{Network Architecture of \MethodName{}}
\label{sec:our_method}
The network architecture of \MethodName{} comprises a tone adjustment network and a \OurHypernet{} as shown in \Fig{\ref{fig:fig3_framework}}. 
The tone adjustment network, for which we adopt an existing image enhancement network, takes an input image and alters the tonal properties of the image, while the \OurHypernet{} takes a target text description as input and modulates the parameters of the tone adjustment network for text-based tone adjustment.
Our design philosophy can be regarded as modulating a backbone network using a hyper-network, a strategy whose efficacy has been demonstrated in various prior studies~\cite{ratzlaff2019hypergan, mahabadi2021parameter, alaluf2022hyperstyle, dynagan}.
Here, the selection of the backbone network and defining an efficient and effective modulation strategy are crucial, and we describe our choices in the following.

\paragraph{Tone Adjustment Network.}
For the tone adjustment network, we adopt the image enhancement network of Yang~\etal~\cite{yang2022adaint}, which is based on 3D LUTs.
Here, we take a brief look at Yang~\etal's method for completeness. 
Their method takes an input image and extracts a feature map using a convolutional neural network (CNN).
Then, from the feature map, a weight predictor network estimates the weights $w$ of learnable basis LUTs.
The basis 3D LUTs are then linearly combined using the weights $w$ to form a fused 3D LUT.
Meanwhile, from the feature map extracted by the CNN, an AdaInt module estimates sampling coordinates, which determine the source color values of the fused 3D LUT.
Specifically, an AdaInt module estimates sampling coordinates $x_c(i)$ where $c \in \{r, g, b\}$ and $i$ denote the color axis and an index along each color axis.
For example, $x_r(1)=0.2$, $x_g(2)=0.3$, and $x_b(3)=0.4$ mean that the color value $(0.2, 0.3, 0.4)$ of an input image will be converted to the color value of the fused 3D LUT at $(1,2,3)$.
Then, the non-uniform sampling step resamples the fused 3D LUT according to the estimated sampling coordinates, and obtains a final image-adaptive 3D LUT.
Finally, the color values of the input image undergo trilinear-interpolated look-up operations to produce a color-enhanced image.
We refer the readers to \cite{yang2022adaint} for more details.

We build our approach on top of the method of Yang~\etal~\cite{yang2022adaint} because it adjusts only the tonal properties of an input image using LUTs without modifying the contents, and provides an efficient and effective architecture leveraging basis LUTs.
In our approach, we use the basis LUTs from the Yang \etal's model, which are pre-trained for the image enhancement task, without modification.
We modulate only the parameters of the weight predictor and the AdaInt module with our \OurHypernet{} to construct an \emph{image-text adaptive 3D LUT} as shown in \Fig{\ref{fig:fig3_framework}}.
These two modules are the most crucial ones as they determine how to combine the basis 3D LUTs, how to resample a fused 3D LUT, and ultimately how to adjust an image.
All the other parameters including the basis LUTs are kept frozen.
This selective modulation enables the efficient and effective learning of our \OurHypernet{}.

\paragraph{Text Adapter.}
The text adapter modulates the parameters for the AdaInt and weight predictor modules according to a target text description $\hat{T}$.
To this end, the text adapter first embeds the input target text description $\hat{T}$ into the CLIP embedding space using a CLIP text encoder $E_T$.
The text adapter also uses a source text description $T$, which is fixed to ``normal photo'', to represent the tone of the input image.
To embed $T$ and $\hat{T}$, we adopt the CLIP-RN50 text encoder.
Then, it computes a directional vector by computing the difference between the embeddings of the target and source text descriptions, i.e., $E_T(\hat{T})-E_T(T)$,
and feeds the directional vector to a two-layered MLP to compute modulation parameter $\Delta\theta$.
Finally, similar to the previous hyper-network-based approach~\cite{alaluf2022hyperstyle}, the text adapter modulates the parameters of the AdaInt and weight predictor modules as:
\begin{equation}
    \label{eq:modulation_of_parameter}
    \hat{\theta} = \theta \otimes (1+s\cdot\Delta\theta)
\end{equation}
where $\theta$ is the original parameter of the AdaInt module and the weight predictor, and $\hat{\theta}$ is their modulated parameter.
$s$ is a scaling factor that controls the degree of adjustment, and $\otimes$ denotes Hadamard product.

Regarding the source description, we have explored different schemes: text-dependent, image-dependent, and neutral schemes.
The text-dependent scheme derives source descriptions from input texts, \eg, antonyms of input texts, while the image-dependent scheme utilizes the CLIP image embedding of an input image instead of the CLIP text embedding of a source text description. The neutral scheme utilizes a fixed text description representing neutrality, such as ``normal photo''.

Among these options, we found that the neutral scheme works the best due to the following reasons.
(i) The image-dependent scheme tends to cause unstable training due to the continuous variation in the source prompt. (ii) The text-dependent scheme is impractical as it requires finding antonyms for the input text, which can be challenging. (iii) The neutral scheme fits our scenario the best, where an image may already be colorful, but a user may still want to edit the image with the keyword ``colorful'' for a more enhanced result. Additionally, we have evaluated different neutral descriptions, \eg, ``ordinary photo'' and ``photo'', and empirically found that ``normal photo'' yields the best results.

\section{Unsupervised Learning for \MethodName{}.}
\paragraph{Training Strategy.}
We train \MethodName{} on two datasets: one for text descriptions
and the other for source images. We note that those are neither paired nor labeled, i.e., unsupervised learning.
Specifically, 
for text descriptions, we adopt the Color Names Dataset~\cite{colornames} encompassing 30,138 color names with duplicates to learn adjustments to various tones.
Before we feed the descriptions to the CLIP text encoder, we append ``photo'' to each description (e.g., a color name ``red'' is transformed into ``red photo'').
For source images, we use 4,500 source images from the MIT-Adobe 5K dataset~\cite{ma5k}.
We augment the images with random-crop and horizontal-flip, and random adjustments of bright and saturation, following previous work~\cite{zeng2020learning}.
For initialization of the \OurHypernet{}, the weights are sampled from a uniform distribution $\mathcal{U}$(0, 0.01), and the biases are set to zero.
For training, we use the Adam optimizer~\cite{Adam} 
with the learning rate $1e-3$ and the coefficients $\beta_1 = 0.9$ and $\beta_2 = 0.999$.

\paragraph{Training Loss.}
Given a randomly sampled pair of a target description $\hat{T}$ and a source image $I$,
we train \MethodName{} using a loss $\mathcal{L}$ defined as:
\begin{equation}
    \mathcal{L} = \lambda_{\text{content}}\mathcal{L}_{\text{content}} + \lambda_{\text{CLIP}}\mathcal{L}_{\text{CLIP}} + \lambda_{\text{LUT}}\mathcal{L}_{\text{LUT}} \;, \label{eq:total-loss}
\end{equation} 
where $\mathcal{L}_\text{content}$, $\mathcal{L}_\text{CLIP}$, and $\mathcal{L}_\text{LUT}$ represent a content-preserving loss, a CLIP directional loss, and an LUT regularization loss, respectively. The parameters $\lambda_{\text{content}}$, $\lambda_{\text{CLIP}}$, and $\lambda_{\text{LUT}}$ are used to balance these loss terms, and we set each of them to $1$.

The content-preserving loss $\mathcal{L}_{\text{content}}$ prevents the tone-adjusted output $\hat{I}$ from excessively deviating from the source image $I$, and is defined as the mean-squared error (MSE) between $I$ and $\hat{I}$.
The CLIP directional loss $\mathcal{L}_\textrm{CLIP}$ aligns the difference from $I$ to $\hat{I}$ with that from $T$ to $\hat{T}$.
To this end, we employ the CLIP directional loss of StyleGAN-NADA~\cite{gal2022stylegan}, which is defined as:
\begin{align}
    \mathcal{L}_{\text{CLIP}}(\Delta{I}, \Delta{T}) &= 1 - \frac{\Delta{I}\cdot\Delta{T}}{\|\Delta{I}\| \|\Delta{T}\|} \;
\end{align}
where $\Delta{I} := E_I(\hat{I}) - E_I(I)$ and $\Delta{T} := E_T(\hat{T}) - E_T(T)$.

Lastly, we devise the LUT regularization loss
$\mathcal{L}_{\text{LUT}} = \lambda_{\text{weight}}\mathcal{L}_{\text{weight}} + \lambda_{\text{interval}}\mathcal{L}_{\text{interval}}$ 
to ensure that the image-text adaptive 3D LUT is locally smooth, thereby enabling natural and stable adjustments.
We set $\lambda_{\text{weight}}$ and $\lambda_{\text{interval}}$ to $1e-4$ and $0.5$, respectively.
The regularization is given in two-fold: $\mathcal{L}_{\text{weight}}$ and $\mathcal{L}_{\text{interval}}$ for the weight predictor module 
and for the AdaInt module, respectively.
We use L2 regularization $\mathcal{L}_{\text{weight}} := \|w\|^2$ on the weights $w$ of the basis LUTs to discourage individual weights from becoming disproportionately large compared to the others.

$\mathcal{L}_{\text{interval}}$ is a sampling interval loss for regularizing the adaptive sampling coordinates from the AdaInt module, which is defined as:
\vspace{-0.2cm}
\begin{align}
    \!\!\mathcal{L}_{\text{interval}} \!= \!\!\! \sum_{l, i, j, k, c} &\left| \frac{L_c^{(l)}{(i + 1, j, k)} - L_c^{(l)}{(i, j, k)}}{(x_r({i+1}) - x_r({i}))^\alpha} \right|^2 \nonumber \\
 + &\left| \frac{L_c^{(l)}(i, j + 1, k) - L_c^{(l)}(i, j, k)}{(x_g(j+1) - x_g(j))^\alpha} \right|^2 
 \nonumber \\
 + &\left| \frac{L_c^{(l)}(i, j, k + 1) - L_c^{(l)}(i, j, k)}{(x_b(k+1) - x_b(k))^\alpha} \right|^2 \!\!,
\end{align}
where $l\in\{1,\cdots,L\}$ is an index to a basis LUT, and $L$ is the number of the basis LUTs, which is $3$ in our framework following Yang \etal.'s method~\cite{yang2022adaint}.
$c \in\{r,g,b\}$ is an index to a color channel.
$L_c^{(l)}(i,j,k)$ is the color value of the $c$ color channel of the $l$-th basis 3D LUT at $(i,j,k)$.
$x_{c}(i)$ is the $i$-th sampling coordinate for the $c$ color channel estimated by the AdaInt module.
$\alpha$ is a hyper-parameter to control the regularization strength.
Larger $\alpha$ more strongly penalizes narrow intervals between consecutive sampling coordinates, 
leading to more evenly distributed sampling coordinates, so that disrupt color changes can be avoided.
Conversely, smaller $\alpha$ gives a weaker penalty to narrow intervals, allowing the sampling coordinates to better reflect the target text description.
In our experiments, we set $\alpha$ to $0.7$.
$\mathcal{L}_{\text{interval}}$ ensures that the distances in the estimated sampling coordinates, i.e., the sampling intervals, are not too small especially when neighboring LUT entries have large color difference, 
thereby preventing abrupt color changes in tone-adjusted images.
\begin{figure*}[t!]
    \centerline{\scalebox{0.95}{\includegraphics[width=0.95\textwidth]{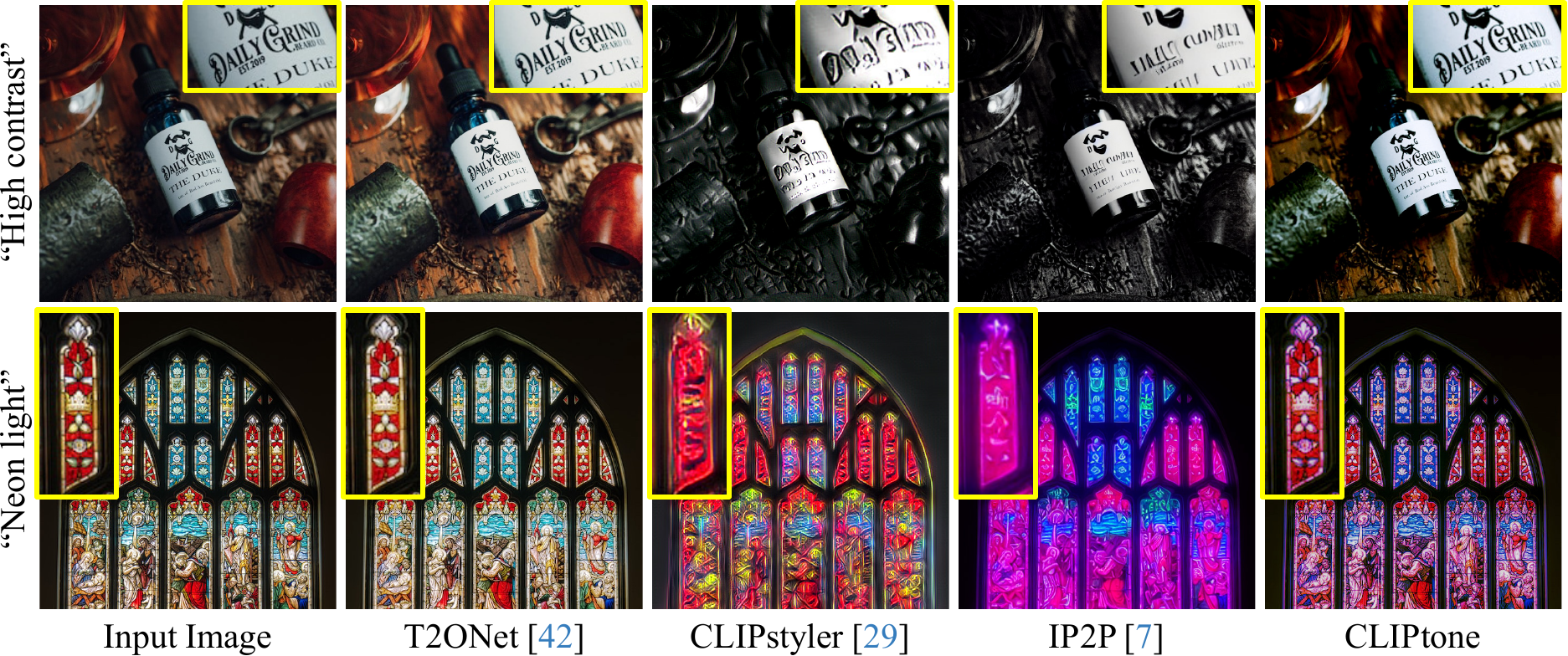}}}
    \caption{
    Qualitative comparisons against baselines for modifications of input images according to given text descriptions.
    T2ONet~\cite{t2onet} induces only subtle changes and fails to perform appropriate adjustments.
    While CLIPstyler~\cite{kwon2022clipstyler} and IP2P~\cite{brooks2023instructpix2pix} make appropriate adjustments aligned with the text descriptions, they fail to preserve the image contents. In contrast, \MethodName{} successfully makes appropriate adjustments while preserving the contents of the input images.
    }
    \label{fig:fig4_qualitative}
    \vspace{-0.4cm}
\end{figure*}
\section{Experiments}
\label{sec:experiments}
\subsection{Comparative Evaluation}
\paragraph{Baselines.}
As \MethodName{} is the first unsupervised learning-based approach for text-based image tone adjustment, we compare it with other state-of-the-art text-based image manipulation methods:
T2ONet~\cite{t2onet}, IP2P~\cite{brooks2023instructpix2pix}, and CLIPstyler~\cite{kwon2022clipstyler}.
T2ONet~\cite{t2onet} is a request-based image enhancement model, which is trained on text-based requests and corresponding before-and-after images in a supervised manner.
IP2P~\cite{brooks2023instructpix2pix} is also a request-based image editing model that fine-tunes a pretrained text-based image generative model to accommodate a source image and a text-based request.
CLIPstyler~\cite{kwon2022clipstyler} is a style transfer model, which is trained to increase the directional CLIP similarity for a single target text description.
As these models use different forms for target descriptions, we alter input text descriptions as follows:
A text description ``krypton'' is transformed to ``krypton photo'' for \MethodName{}, ``Make a krypton photo.'' for T2ONet~\cite{t2onet} and IP2P~\cite{brooks2023instructpix2pix}, and ``krypton'' for CLIPstyler~\cite{kwon2022clipstyler}.
For the baseline models, we use the authors' official implementations.

\paragraph{Qualitative Comparisons.}
\Fig{\ref{fig:fig4_qualitative}} shows a qualitative comparison. In the figure, T2ONet~\cite{t2onet} barely changes the input images and fails to perform appropriate adjustment.
Although CLIPstyler~\cite{kwon2022clipstyler} and IP2P~\cite{brooks2023instructpix2pix} succeed in making adjustment to fit the descriptions, they fail to preserve the structures of the input images. 
\MethodName{} outperforms all the other methods, achieving both appropriate adjustments reflecting the target descriptions and successful preservation of the structural information in the input images.

\begin{table}[t]
\centering
\scalebox{0.78}{
\begin{tabular}{c|c|c|c}
\hline
\Xhline{3\arrayrulewidth}
Method &
  \begin{tabular}[c]{@{}c@{}}Grayscale       \\ SSIM $\uparrow$         \end{tabular} & 
  \begin{tabular}[c]{@{}c@{}}CLIP image      \\ similarity $\uparrow$        \end{tabular} &
  \begin{tabular}[c]{@{}c@{}}CLIP text-image \\ directional similarity $\uparrow$  \end{tabular}  \\ \hline
T2ONet~\cite{t2onet}          & \textbf{0.936} & \textbf{0.994} & 0.021 \\
CLIPstyler~\cite{kwon2022clipstyler}     & 0.484 & 0.669  & \textbf{0.095} \\
IP2P~\cite{brooks2023instructpix2pix} & 0.765  & 0.932 & 0.073  \\
CLIPtone            & \underline {0.858} & \underline {0.975} & \underline {0.083} \\ \hline
\Xhline{3\arrayrulewidth}
\end{tabular}}
    \vspace{-0.2cm}
\caption{Quantitative comparison of different methods on 500 images from the MIT-Adobe 5K dataset~\cite{ma5k} and manually collected 50 tone-related text descriptions. 
\MethodName{} achieves overall higher scores in both Grayscale SSIM and CLIP image similarity, which evaluate structural preservation, as well as in CLIP text-image directional similarity, which assesses the alignment with text descriptions, compared to existing methods.}
\vspace{-0.2cm}
\label{tab:tbl1_quantitative_comparison}
\end{table}

\paragraph{Quantitative Comparisons.}
\label{sec:quantitative}
For quantitative evaluation, we construct a test set consisting of 500 source images from MIT-Adobe 5K~\cite{ma5k}, which are not used for training \MethodName{}, and 50 manually collected text descriptions related to tonal properties such as ``warm'', ``cold'', and ``pastel tone''.
We provide all the descriptions used in our evaluation in the supplementary material.

To assess the adjustments by each method, we focus on two primary factors: 
the preservation of image structures, and the alignment with text descriptions.
To assess the structure preservation, we use grayscale SSIM and CLIP image similarity. Grayscale SSIM computes the SSIM index between two images after grayscale conversion, while CLIP image similarity measures the cosine similarity between two images in the CLIP embedding space to assess the similarity between their contents. 

However, it is important to note that both metrics are limited in assessing the structure preservation since assessing the structure preservation between two images is not a trivial task, which requires isolating the image structure from tones and colors. Nevertheless, both metrics are less affected by tones and colors than other metrics such as PSNR since grayscale SSIM discards the colors and also includes contrast normalization, and CLIP image similarity focuses on content similarity rather than tones and colors.

To assess alignment with text descriptions, we measure the CLIP text-image directional similarity.
Specifically, given a pair of an input image and a target text description, we compute the directional vector between the CLIP embeddings of the input image and its tone-adjusted image, and the directional vector between the CLIP embeddings of the source and target text descriptions.
Then, we compute the cosine similarity between the directional vectors.
Meanwhile, as T2ONet~\cite{t2onet} supports only predefined descriptions for adjustments, we evaluate the performance of T2ONet only on 21 compatible descriptions out of 50.

In \Tbl{\ref{tab:tbl1_quantitative_comparison}}, we report the quantitative comparison.
The CLIP image similarity and CLIP text-image directional similarity scores in the table are computed using the CLIP ViT-B/16 model, which is not used for training any methods for a fair comparison.
Other results with other CLIP models are included in the supplementary material.
In the table, T2ONet~\cite{t2onet}, which induces subtle modifications in adjustment process, yields high scores in Grayscale SSIM and CLIP image similarity. Yet, it markedly underperforms in CLIP text-image directional similarity than the other methods.
Since CLIPstyler~\cite{kwon2022clipstyler} is trained solely on target descriptions, it records the highest CLIP text-image directional similarity score. 
However, it struggles to preserve image structures, leading to the lowest scores in Grayscale SSIM and CLIP image similarity.
\MethodName{} achieves high scores in all three metrics, demonstrating that it outperforms existing methods when considering both primary factors.

\begin{figure*}[t]
    \centerline{\includegraphics[width=0.88\textwidth]{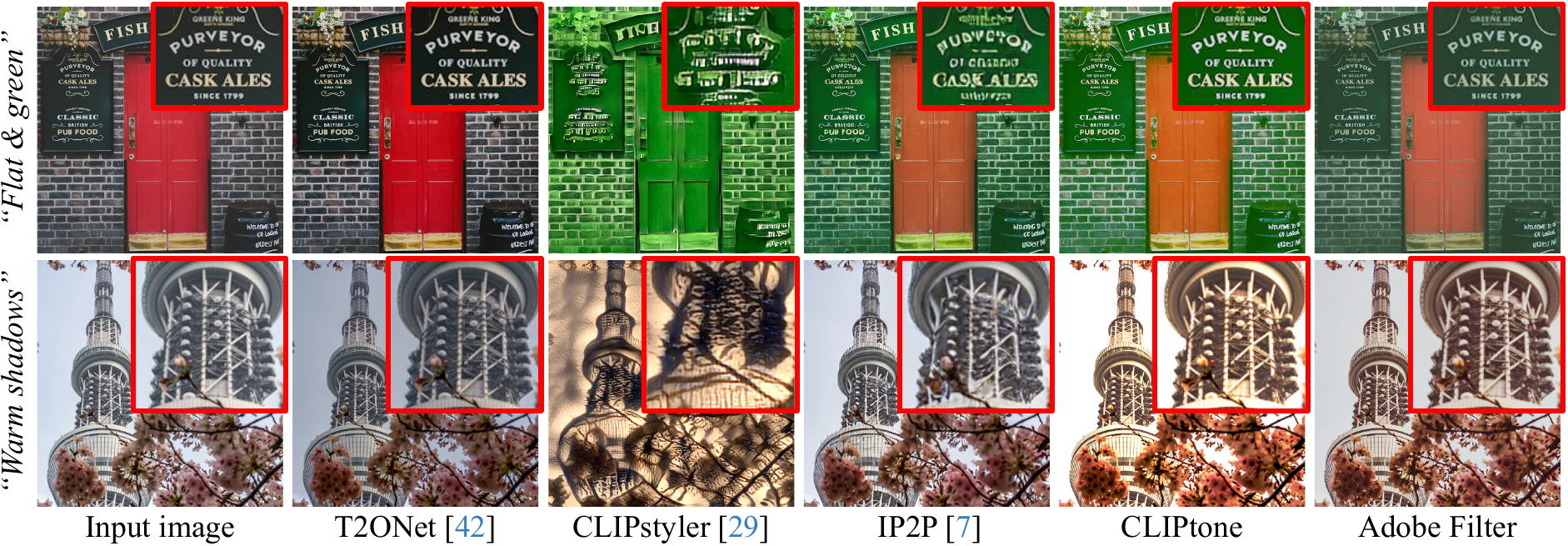}}
    \vspace{-0.2cm}
    \caption{
    Qualitative comparison against pseudo-ground-truth images created by tone adjustment filters of Adobe Lightroom Classic~\cite{adobelightroomclassic}.
    Compared to the baselines, 
    \MethodName{} performs more similar adjustment to the Adobe filters while preserving the original image structure.
    }
    \label{fig:fig10_crafted_filters}
    \vspace{-0.6cm}
\end{figure*}
\paragraph{Comparison using Pseudo-Ground-Truth Images.}
We also compare the baseline methods and ours using pseudo-ground-truth images, which are generated by applying nine tone adjustment filters of Adobe Lightroom Classic \cite{adobelightroomclassic} to our test set as done in \cref{sec:method}.
\cref{fig:fig10_crafted_filters} shows a qualitative comparison.
As the figure shows, \MethodName{} excels not only in preserving image structures but also in adjusting tones, producing results more similar to the results of the filters designed by experts.
For quantitative analysis, we measure the structural similarity between tone-adjusted images and their corresponding pseudo-ground-truth images using CLIP image similarity.
We also measure the similarity between the adjustments made to a tone-adjusted image and to its corresponding pseudo-ground-truth image for a given input image using CLIP directional similarity.
\cref{tab:tbl4_crafted_filters} shows that \MethodName{} achieves the highest CLIP directional similarity score and the second highest CLIP image similarity score, indicating that \MethodName{} performs tone adjustment similar to the filters manually designed by experts.

\begin{figure}[t]
  \centering
    \includegraphics[width=0.85\linewidth]{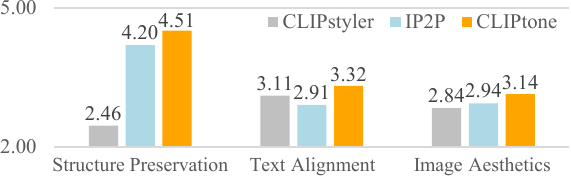}
    \vspace{-0.2cm}
   \caption{Summary of the user study with 20 participants evaluating 30 examples.
Each graph represents the average score of each of the compared methods for each of the following questions:
(a) How well the structure of the input image is preserved.
(b) How well the modification suits the text description.
(c) The aesthetic quality of the result image.}
   \label{fig:fig5_user_study}
   \vspace{-0.4cm}
\end{figure}
\begin{table}[t]
\centering
\scalebox{0.72}{
\begin{tabular}{c|c|c|c|c}
\hline
\Xhline{3\arrayrulewidth}
 Methods  & T2ONet~\cite{t2onet} & CLIPstyler~\cite{kwon2022clipstyler} & IP2P~\cite{brooks2023instructpix2pix} & CLIPtone \\
 \hline
\begin{tabular}[c]{@{}c@{}}CLIP image\\similarity \end{tabular} & $\textbf{0.956}$ & 0.704 & 0.891 & \underline{0.948} \\
\hline
 \begin{tabular}[c]{@{}c@{}}CLIP directional\\similarity \end{tabular} &\underline{0.359} & 0.297 & 0.310 & $\textbf{0.467}$ \\ \hline
\Xhline{3\arrayrulewidth}
\end{tabular}}
\vspace{-0.2cm}
\caption{Quantitative comparison using pseudo-ground-truth images created by tone adjustment filters of Adobe Lightroom Classic~\cite{adobelightroomclassic}.}
\label{tab:tbl4_crafted_filters}
\vspace{-0.4cm}
\end{table}
\begin{figure}[t]
  \centering
    \includegraphics[width=0.88\linewidth]{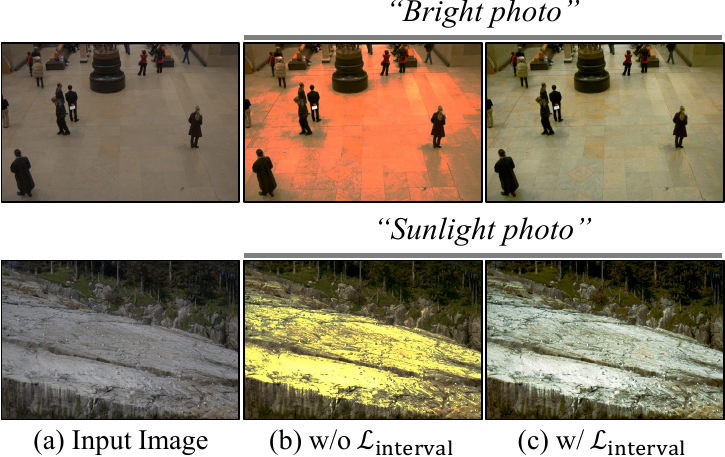}
    \vspace{-0.1cm}
   \caption{
   The effect of the sampling interval loss $\mathcal{L}_{\text{interval}}$.
   Without the loss, \MethodName{} may perform excessive adjustments solely to align with the text descriptions, compromising aesthetic quality.
   }
   \vspace{-0.5cm}
   \label{fig:fig6_ablation}

\end{figure}

\begin{table}[t]
\centering
\scalebox{0.78}{
\begin{tabular}{c|c|c|c|c}
\hline
\Xhline{3\arrayrulewidth}
Method & T2ONet~\cite{t2onet} & CLIPstyler~\cite{kwon2022clipstyler} & IP2P~\cite{brooks2023instructpix2pix} & \MethodName{} \\ \hline
Param (M) & \underline {21.1} & \textbf {6.7} & 1016.9 & 111.4 \\
Time (ms) & 45.1 & \underline {26.6} & 7177.4 & \textbf{10.3} \\ \hline
\Xhline{3\arrayrulewidth}
\end{tabular}}
    \vspace{-0.2cm}
\caption{
Comparison of the number of parameters (M) and inference times (ms) for processing a $512\times512$-sized image using an NVIDIA GeForce RTX 3090 GPU. 
}
\vspace{-0.4cm}
\label{tab:tbl2_infer_time_comparison}
\end{table}

\begin{table}[t]
\centering
\scalebox{0.68}{
\begin{tabular}{c|c|c|c}
\hline
\Xhline{3\arrayrulewidth}
Method &
  \begin{tabular}[c]{@{}c@{}}Grayscale       \\ SSIM $\uparrow$              \end{tabular} & 
  \begin{tabular}[c]{@{}c@{}}CLIP Image      \\ Similarity $\uparrow$        \end{tabular} &
  \begin{tabular}[c]{@{}c@{}}CLIP Text-Image \\ Directional Similarity $\uparrow$ \end{tabular} \\ \hline

\begin{tabular}[c]{@{}c@{}}CLIPtone-0\\(trained w/o target keyword) \end{tabular} & \textbf {0.859} & \textbf {0.968} & \underline {0.087}  \\ \hline
\begin{tabular}[c]{@{}c@{}}\MethodName{}\\(trained w/ target keyword) \end{tabular} & \underline {0.858} & \underline {0.964} & \textbf {0.089}  \\ \hline
\Xhline{3\arrayrulewidth}
\end{tabular}}
\vspace{-0.2cm}
\caption{
Quantitative comparison of \MethodName{}-0 trained without target keywords and \MethodName{} trained with target keywords.
\MethodName{}-0 performs nearly as well as \MethodName{} in all the metrics.
}
\vspace{-0.4cm}
\label{tab:tbl3_zeroshot}
\end{table}

\begin{figure}[t]
  \centering
    \includegraphics[width=0.88\linewidth]{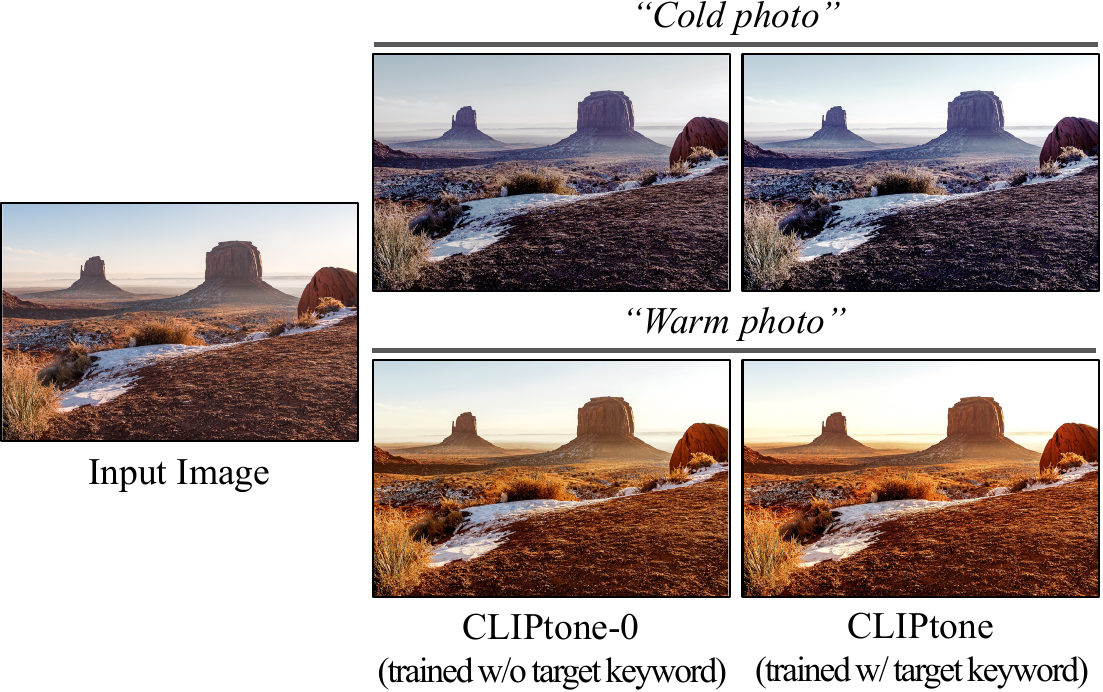}
    \vspace{-0.2cm}
   \caption{
   Qualitative comparison of \MethodName{}-0 trained without target keywords and \MethodName{} trained with target keywords.
   In both cases, both models make appropriate and similar adjustments implying that \MethodName{} is capable of robustly adjusting even to novel descriptions not encountered during training.
   }
   \label{fig:fig7_zeroshot}
   \vspace{-0.3cm}
\end{figure}

\begin{figure}[t]
  \centering
    \includegraphics[width=0.88\linewidth]{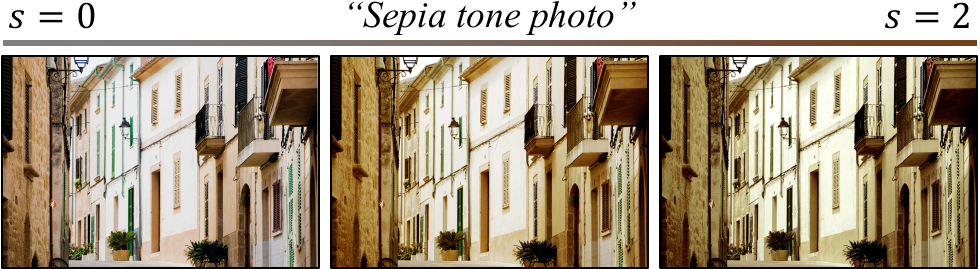}
    \vspace{-0.2cm}
   \caption{Controlling the strength of adjustment by adjusting the modulation scaling factor $s \in \{0, 1, 2\}$. A result with the default scaling factor $s=1$ is shown in the middle.
}
   \label{fig:fig9_control_strength}
      \vspace{-0.6cm}
\end{figure}
\paragraph{User Study.}
Given that image tone adjustment is fundamentally a human-centric task, we conducted a user study to evaluate the baseline methods and ours.
For this study, we recruited 20 participants from our institution. 
Each participant was presented 30 examples consisting of an input image and randomly arranged resulting images from each method, and was asked to score these on a scale from 1 (poor) to 5 (excellent) based on the following criteria:
(a) How well the structure of the input image is preserved.
(b) How well the modification suits the text description.
(c) The aesthetic quality of the result image.
The complete questionnaires are included in the supplementary material.
\Fig{\ref{fig:fig5_user_study}} illustrates the average of the results of the user study.
Against all the other methods, \MethodName{} is preferred in all three aspects, demonstrating that \MethodName{} outperforms existing methods from the perspective of human perception.

\paragraph{Model Complexity.}
As shown in \Tbl{\ref{tab:tbl2_infer_time_comparison}},
\MethodName{} has more parameters than T2ONet~\cite{t2onet} and CLIPstyler~\cite{kwon2022clipstyler}, each designed to address single and restricted text descriptions, respectively.
Nevertheless, \MethodName{} exhibits notably shorter inference times than the others thanks to its efficient backbone network and our hyper-network-based approach.

\subsection{Deeper Analysis of \MethodName{}}
In this section, we conduct a detailed analysis on \MethodName{}, including the validation of the sampling interval loss and a series of experiments that demonstrate the unique and practical advantages of \MethodName{}.

\paragraph{Sampling Interval Loss.}
\Fig{\ref{fig:fig6_ablation}} shows that the absence of $\mathcal{L}_{\text{interval}}$ causes the model to over-adjust, significantly diminishing the aesthetic quality of the output image. This happens because the estimated sampling coordinates from the AdaInt module lead to the construction of an image-text adaptive 3D LUTs with abrupt color transitions, focusing on matching the target text descriptions.
In contrast, \MethodName{} with the sampling interval loss constructs an image-text adaptive 3D LUT with more gradual color transitions, ensuring that the result image is not only aligned with the target text descriptions but also visually appealing.

\paragraph{Zero-shot Prediction.}
Thanks to leveraging CLIP that is pretrained on a vast amount of image-text pairs,
\MethodName{} can also make suitable adjustments for text descriptions unseen during training.
Here we verify this zero-shot prediction capability of \MethodName{}.
To this end, we construct a test set using 50 target keywords used in the quantitative evaluation. We then exclude any text descriptions that include the keywords from our training set.
As a result, we exclude 2,890 text descriptions, and obtain a subset of 27,248 text descriptions.
Then, we train \MethodName{} on the subset. We refer to this version of \MethodName{} as `\MethodName{}-0'.
Finally, we compare \MethodName{}-0 and \MethodName{} on the test set of the sampled target keywords.
\cref{tab:tbl3_zeroshot} and \cref{fig:fig7_zeroshot} present the quantitative and qualitative comparisons.
These comparisons reveal that \MethodName{}-0 performs nearly as well as \MethodName{}, suggesting that \MethodName{} is capable of handling novel text descriptions not included in its training set.
A more comprehensive collection of zero-shot prediction results is included in the supplementary material.

\paragraph{Scaling the Strength of Adjustment.}
\MethodName{} also offers controllable adjustment by changing the scaling factor $s$ in \Eq{\ref{eq:modulation_of_parameter}}, which scales the weight offsets of the backbone network estimated by the \OurHypernet{}.
\Fig{\ref{fig:fig9_control_strength}} illustrates an example that shows smooth transition of the tone according to the scaling factor $s$.
\section{Conclusion}
In this paper, we propose \MethodName{}, the first unsupervised learning-based approach for text-based image tone adjustment.
\MethodName{} effectively extends a 3D LUT-based image enhancement model to accommodate a range of adjustments guided by text descriptions.
Thanks to utilizing CLIP for perceptual supervision, \MethodName{} requires no paired datasets, and also supports zero-shot prediction for unseen text descriptions.
Through comprehensive experiments, we have validated the effectiveness and efficiency of \MethodName{}.

\paragraph{Limitations.}
\MethodName{} has some limitations.
Firstly, the inherent bias in a pretrained CLIP model may lead \MethodName{} to perform adjustments that do not align with human perception.
We include an analysis in the supplementary material.
Secondly, \MethodName{} relies global tone adjustment, thus cannot perform tone adjustment that requires local tone adjustment.
Extending \MethodName{} to support local adjustment could be a promising research avenue.
Thirdly, as \MethodName{} depends solely on the directional vectors from text descriptions, it cannot support adjustments for more diverse and specific styles, particularly those difficult to express with text alone. Leveraging a CLIP image encoder could be an interesting future direction.

\paragraph{Acknowledgements.}
This work was supported by the NRF grant (No.2023R1A2C200494611) and IITP grant (IITP-2021-0-02068, IITP-2019-0-01906) funded by the Korea government (MSIT).

\clearpage
{
    \small
    \bibliographystyle{ieeenat_fullname}
    \bibliography{main}
}

 
\end{document}


\renewcommand{\thesection}{S}
\clearpage

\setcounter{page}{1}
\maketitlesupplementary

\subsection{User Study Details}
\begin{figure*}[t]
    \centerline{\includegraphics[width=\textwidth]{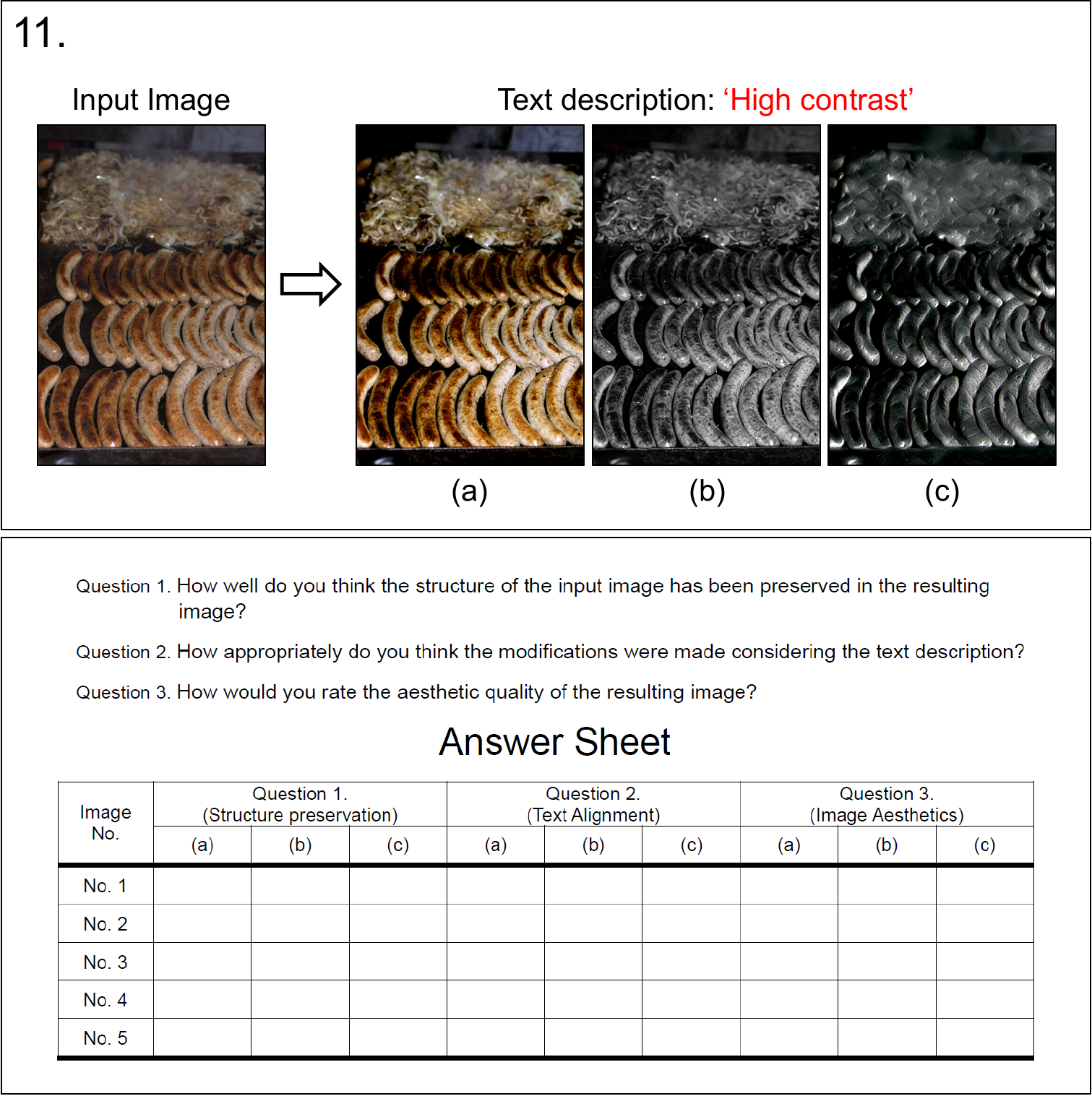}}
    \caption{
    Our user study questionnaire. 
    Each participant was presented upper figure consisting of an input image and results of different methods and asked to complete below answer sheet.
    }
    \label{fig:sup_fig1_user_study}
\end{figure*}

For the user study, we recruited 20 participants from our institution and each participant was asked to complete a questionnaire as shown in \Fig{\ref{fig:sup_fig1_user_study}}.
The questionnaire consists of an answer sheet and examples for randomly selected 30 images and 30 text descriptions from our test set.
In each example, an input image and randomly arranged result images of different methods are displayed.
The compared methods include CLIPstyler~\cite{kwon2022clipstyler}, IP2P~\cite{brooks2023instructpix2pix}, and \MethodName{}, but do not include T2ONet~\cite{t2onet} as it is limited to predefined text descriptions.
Each participant was asked to rate each result on a scale from 1 (poor) to 5 (excellent) based on the following questions: 
\begin{itemize}
\item (Q1) How well do you think the structure of the input image has been preserved in the resulting image?
\item (Q2) How appropriately do you think the modifications were made considering the text description?
\item (Q3) How would you rate the aesthetic quality of the resulting image?
\end{itemize}

\subsection{Detailed Test Set Descriptions}
For quantitative comparison, we manually collected 50 text descriptions that represent tonal properties such as brightness, contrast, color balance, and the overall mood of an image.
The descriptions are as follows: 
\newline
[``\underline{warm}'', ``\underline{bright}'', ``\underline{dark}'', ``\underline{gloomy}'',  ``\underline{bold}'', ``\underline{dull}'', ``\underline{dramatic}'', ``\underline{low contrast}'', ``\underline{colorful}'', ``\underline{vibrant}'', ``\underline{HDR}'', ``\underline{vintage}'', ``\underline{saturated}'', ``\underline{neon light}'', ``\underline{daylight}'', ``\underline{sunlight}'', ``\underline{vivid}'', ``\underline{lively}'', 
``\underline{high contrast}'', ``\underline{faded}'',  ``\underline{beautiful}'',  ``captivating'', ``cozy'', ``cold'', ``monochrome'', ``grayscale'', ``old'', ``modern'', ``cyberpunk'', ``pastel tone'', ``moonlight'', ``nighttime'', ``desaturated'', ``twilight'', ``pop art'',  ``gothic'', ``polaroid'', ``analog'', ``silhouette'', ``cinematic'', ``dreamy'', ``romantic'', ``mysterious'', ``relaxed'', ``dynamic'', ``melancholic'', ``urban'', ``ethereal'', ``matte'', ``edgy'']
\newline
where the 21 underlined words are those usable in T2ONet~\cite{t2onet}.

\subsection{Additional Validation of Hypothesis with Different CLIP Models}
\begin{figure*}[t]
    \centerline{\includegraphics[width=0.95\textwidth]{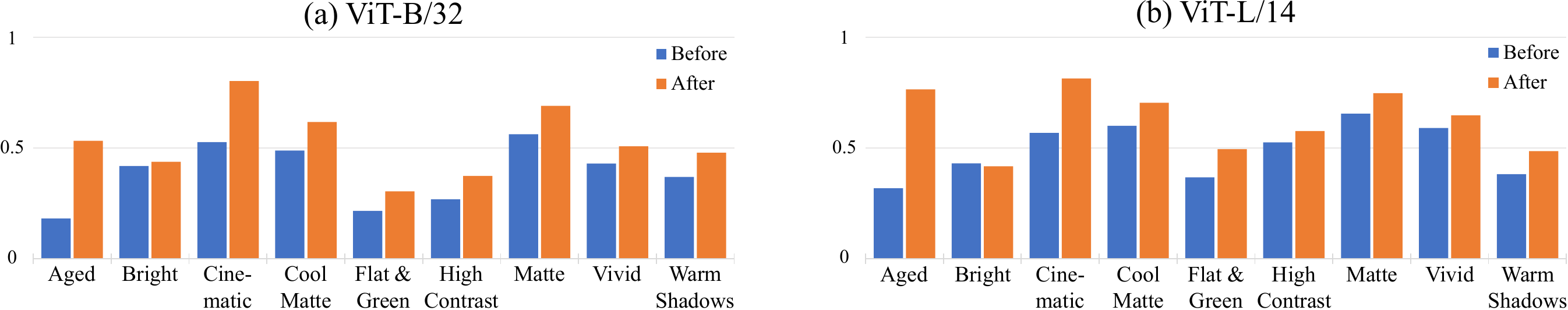}}
    \caption{
    Additional hypothesis validation experiments using different CLIP models.
    Both results consistently show increases in similarities in filtered images, which again implies that CLIP is capable of assessing tonal properties of images, aligning with human perception.
    }
    \label{fig:sup_fig6_clipgraph}
\end{figure*}

\cref{fig:sup_fig6_clipgraph} shows additional hypothesis validation experiments in Sec.~\red{3} in the main paper with different CLIP models: ViT-B/32 and ViT-L/14.
Consistent with the result presented in the main paper, both results demonstrate consistent increases in similarities in filtered images, supporting our hypothesis that CLIP can serve as a perceptual criterion for tone adjustments.

\subsection{Trade-off between Content Loss and CLIP Loss}
\begin{figure*}[t]
    \centerline{\includegraphics[width=1\textwidth]{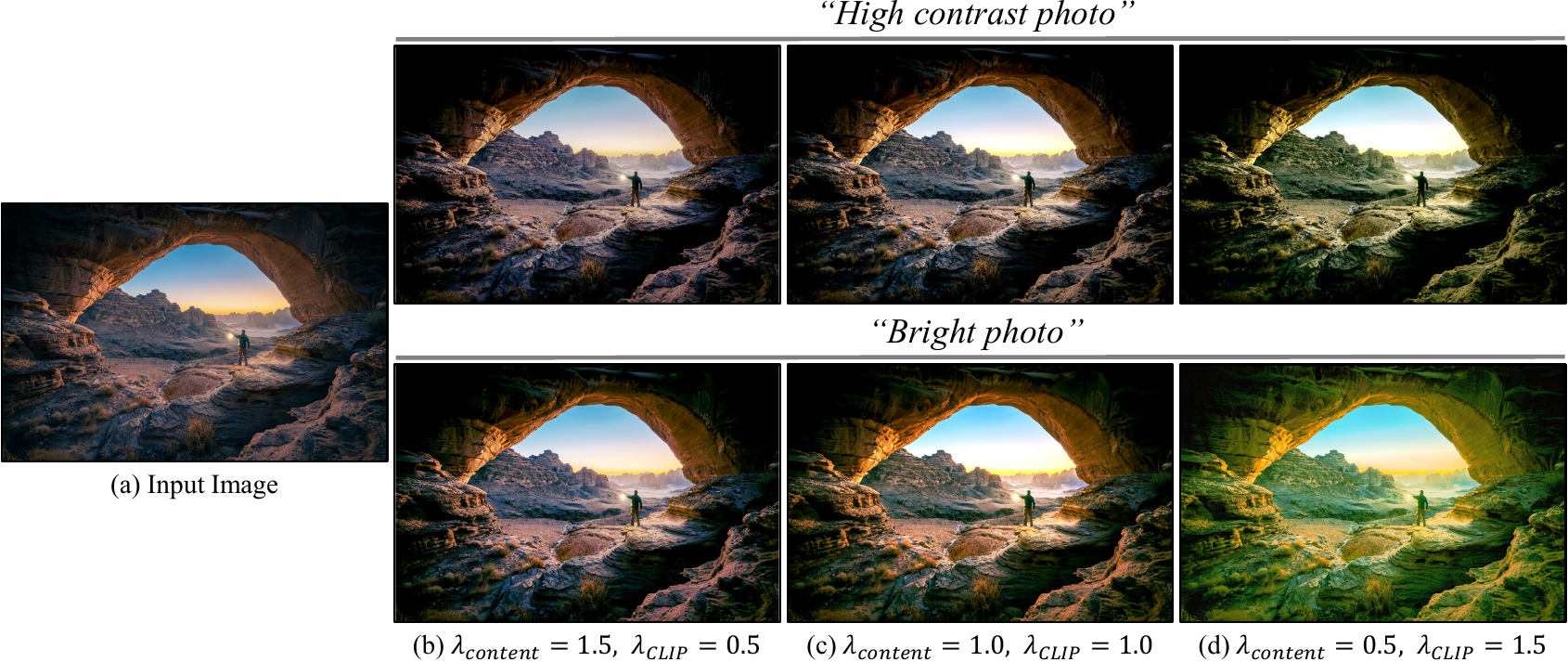}}
    \caption{
    Effects of the content loss $\mathcal{L}_\textrm{content}$ and CLIP loss $\mathcal{L}_\textrm{CLIP}$.
    Larger weight $\lambda_{\text{content}}$ for the content loss leads to tone adjusted results that are closer to the input image,
    while larger weight $\lambda_{\text{content}}$ for the CLIP loss leads to more significant alterations.
    }
    \label{fig:sup_fig7_trade_off}
\end{figure*}

\cref{fig:sup_fig7_trade_off} presents a qualitative example to show the trade-off between the content-preserving loss $\mathcal{L}_{\text{content}}$ and the CLIP loss $\mathcal{L}_{\textrm{CLIP}}$.
As shown in the figure, a larger value for the weight $\lambda_{\text{content}}$ leads to tone adjustment results that are closer to the input image, while a larger value for the weight $\lambda_{\text{CLIP}}$ leads to more significant alterations in the tonal properties of the input image.

\subsection{Color Biases of CLIPtone}
\begin{figure*}[t]
    \centerline{\includegraphics[width=1\textwidth]{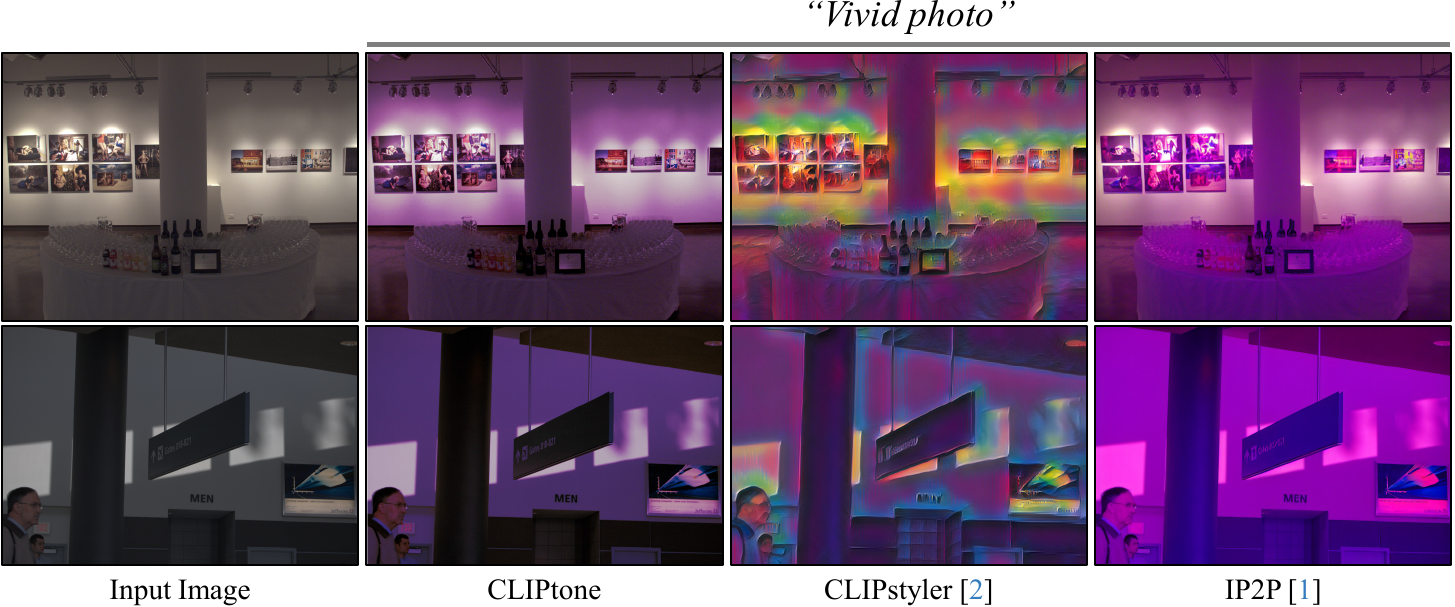}}
    \caption{
    Examples of color bias of CLIP-based methods.
    \MethodName{} and other CLIP-based methods show similar color-biased results that adjust the input image toward purple tone.
    }
    \label{fig:sup_fig8_failure_cases}
\end{figure*}

As mentioned in the main paper, \MethodName{} is not free from limitations. One limitation is that it exhibits unintended color biases for certain text descriptions as shown in \cref{fig:sup_fig8_failure_cases}.
In the figure, \MethodName{} adjusts the input images towards purple for the input text description ``vivid photo''.
Notably, this kind of color bias is not exclusive to our model but is also found in other CLIP-based models, such as CLIPstyler~\cite{kwon2022clipstyler} and IP2P~\cite{brooks2023instructpix2pix}.
This result indicates that such color biases are caused by the inherent biases within the pretrained CLIP models.

\subsection{Additional Qualitative Comparisons}
\begin{figure*}[t]
    \centerline{\includegraphics[width=\textwidth]{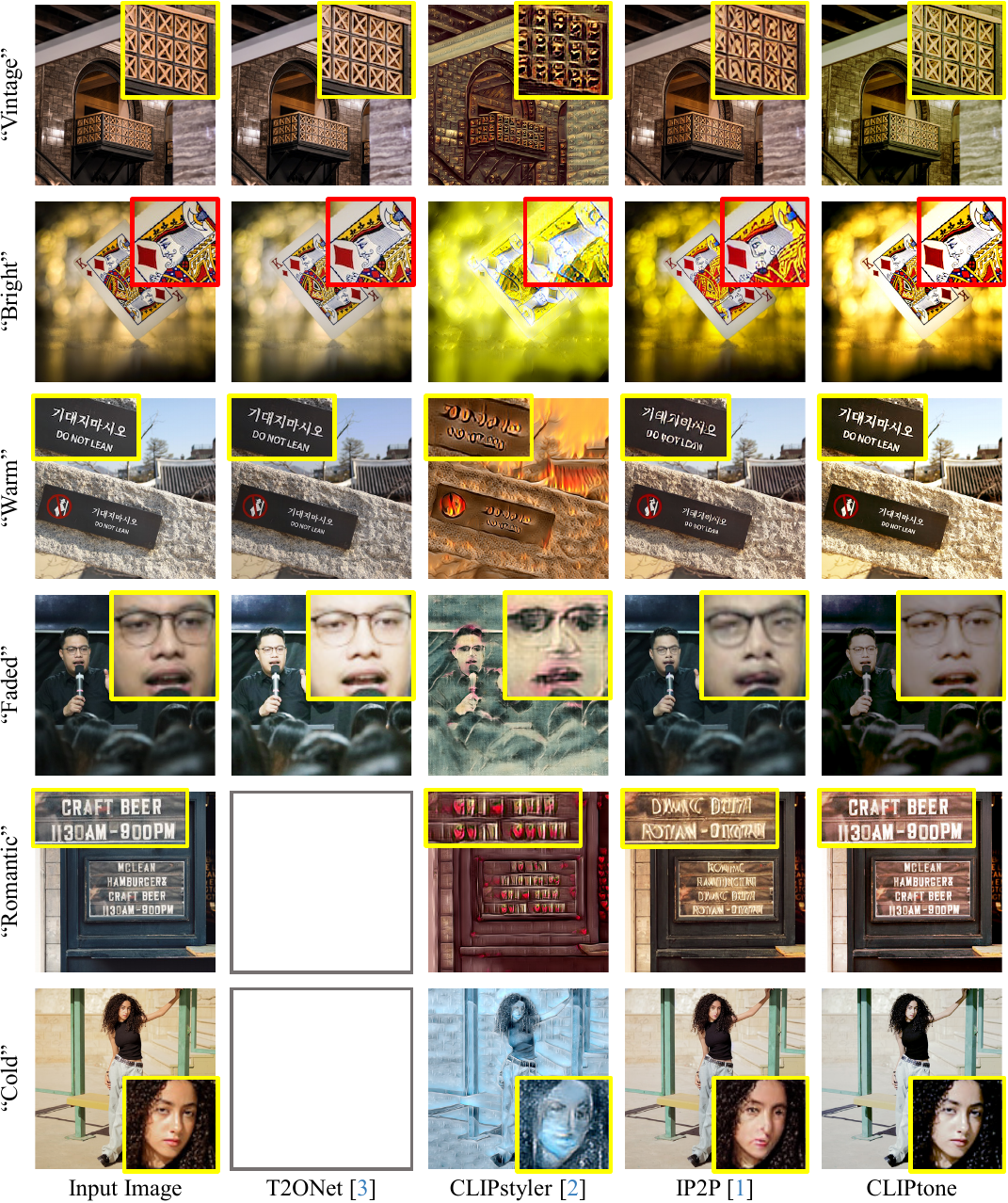}}
    \caption{
    Additional qualitative comparison of different methods.
    T2ONet~\cite{t2onet} does not support ``Romantic'' and ``Cold''.
    }
    \label{fig:sup_fig4_qualitative}
\end{figure*}

\Fig{\ref{fig:sup_fig4_qualitative}} shows additional qualitative comparisons of different methods. 
It highlights that only \MethodName{} successfully makes the appropriate adjustments aligned with the text descriptions while preserving the structure of the image.
Meanwhile, T2ONet~\cite{t2onet} makes only subtle adjustments, such as slight brightness enhancements, failing to adjust adequately according to descriptions.
Both CLIPstyler~\cite{kwon2022clipstyler} and IP2P~\cite{brooks2023instructpix2pix} make adjustments aligned with the descriptions, but struggle to preserve the input image's structure.

\subsection{Additional Quantitative Comparisons with Other Versions of the CLIP Model}
We further report additional quantitative results calculated with other versions of the CLIP model in \Tbl{\ref{tab:sup_tbl1_quantitative_comparison}}.
In the table, RN50 is the version used to train \MethodName{}, ViT-L/14 for IP2P~\cite{brooks2023instructpix2pix}, ViT-B/32 for CLIPstyler~\cite{kwon2022clipstyler}, and RN101 is not used to train any model.
As analyzed in the main paper, T2ONet~\cite{t2onet} generally achieves high scores in CLIP Image Similarity, yet it scores the lowest in CLIP Text-Image Direction Similarity.
In contrast, CLIPstyler~\cite{kwon2022clipstyler} shows an opposite trend in these metrics. 
Both IP2P~\cite{brooks2023instructpix2pix} and \MethodName{} record high scores in both metrics, and among them, \MethodName{} records higher scores.

\subsection{Additional Examples of the Zero-Shot Prediction}
\begin{figure*}[t]
    \centerline{\includegraphics[width=1.1\textwidth]{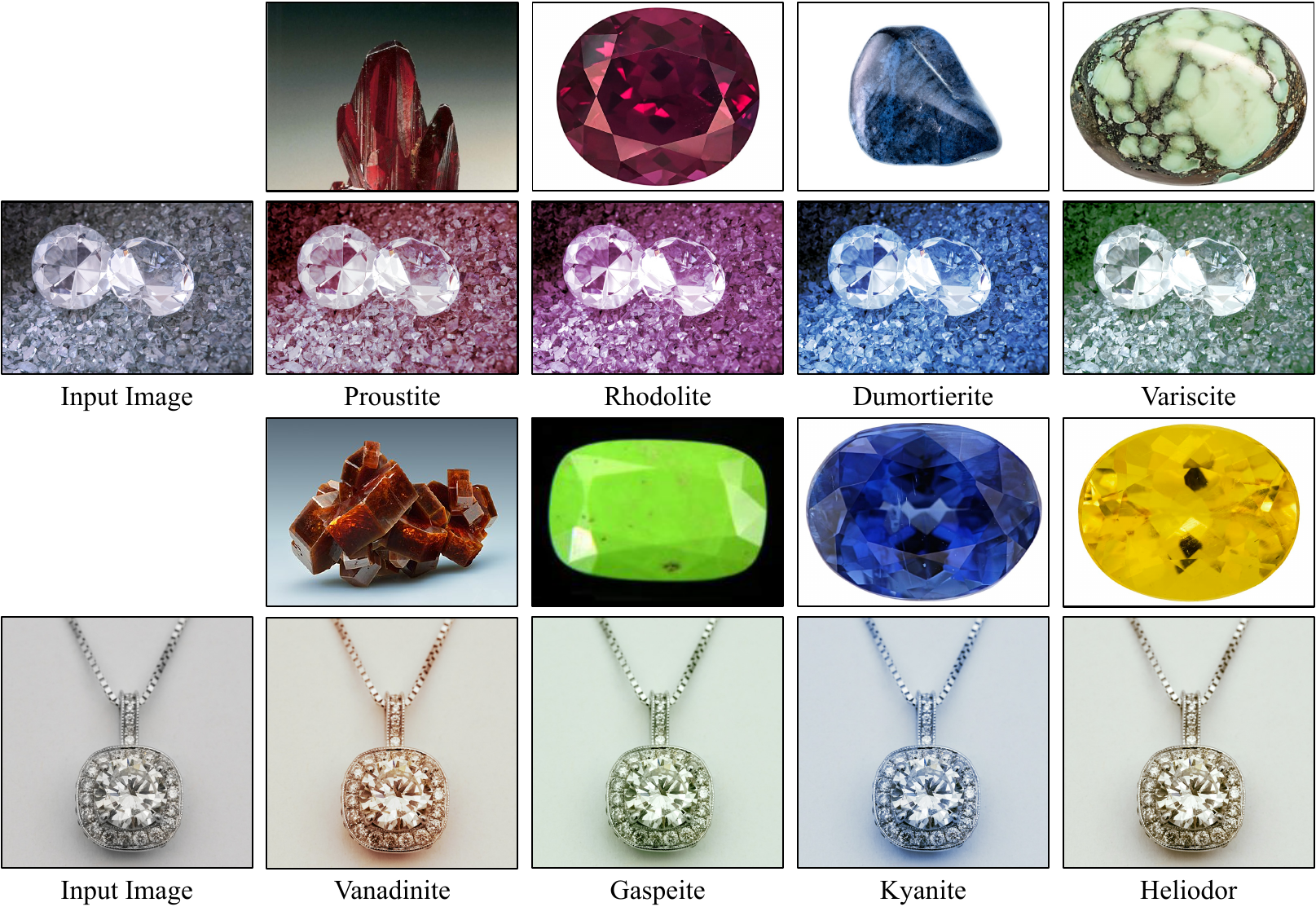}}
    \caption{
    Additional examples of \MethodName{} to demonstrate its zero-shot prediction capability. 
    \MethodName{} successfully executes proper adjustments for words related to minerals, even those not encountered during training. In each example, the top row displays instances of each word to facilitate comprehension.
    }
    \label{fig:sup_fig2_zeroshot}
    \vspace{-3cm}
\end{figure*}

We provide additional examples of the zero-shot prediction in \Fig{\ref{fig:sup_fig2_zeroshot}}.
Although none of the descriptions were included in our training dataset, \MethodName{} successfully captures the unique tonal properties from the text descriptions and makes proper adjustments to input images.

\subsection{Results of \MethodName{} with Long Text Descriptions.}
\begin{figure*}[t]
    \centerline{\includegraphics[width=0.95\textwidth]{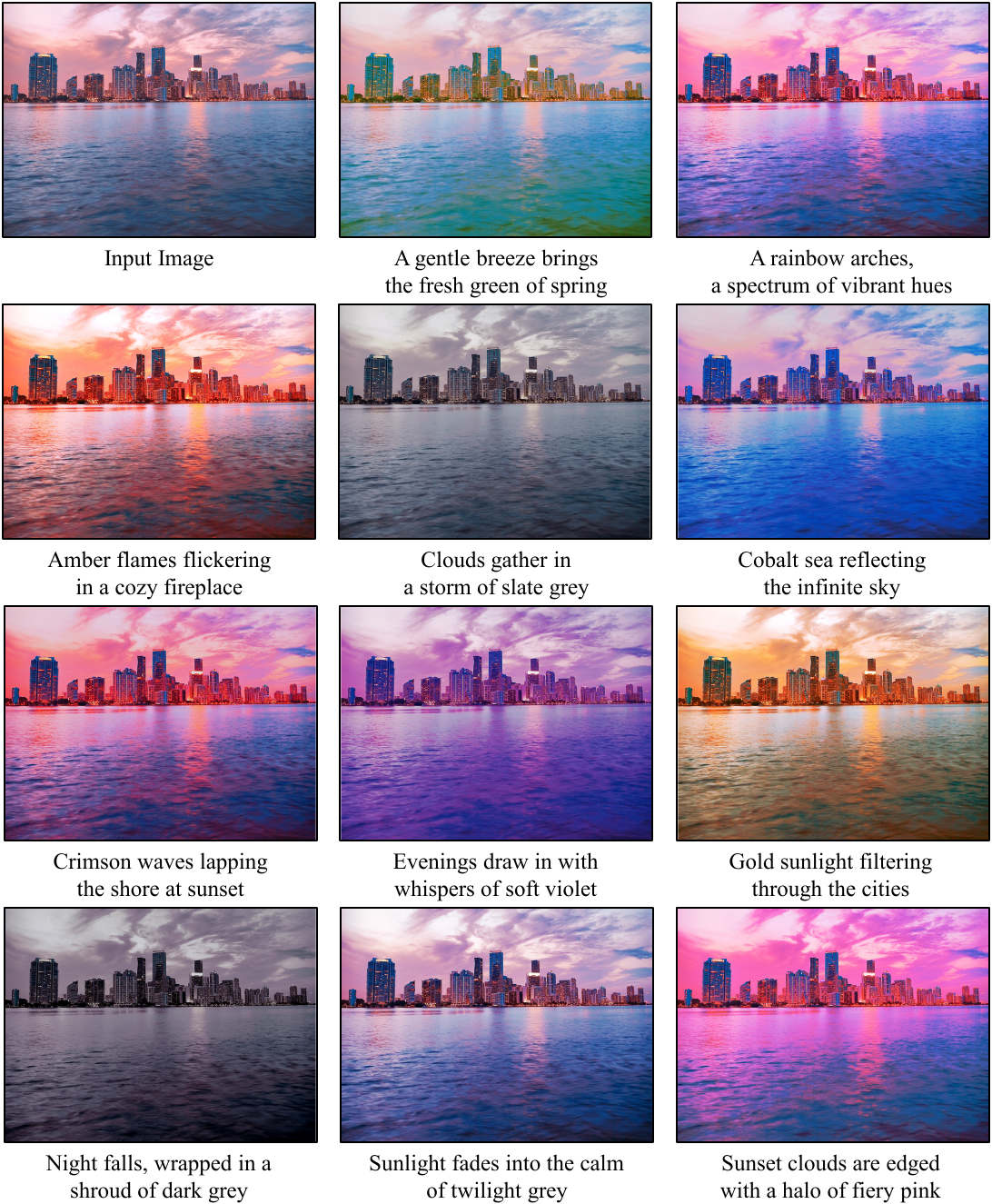}}
    \caption{
    Additional results of \MethodName{} with long and complex text descriptions.
    }
    \label{fig:sup_fig5_long_text}
\end{figure*}

We provide results of \MethodName{} with long text descriptions in \cref{fig:sup_fig5_long_text}. Despite being trained on relatively short descriptions, the results show that \MethodName{} can effectively handle complex and long text descriptions.

\subsection{Additional Results of \MethodName{} with Various Text Descriptions.}
\begin{figure*}[t]
    \centerline{\includegraphics[width=0.95\textwidth]{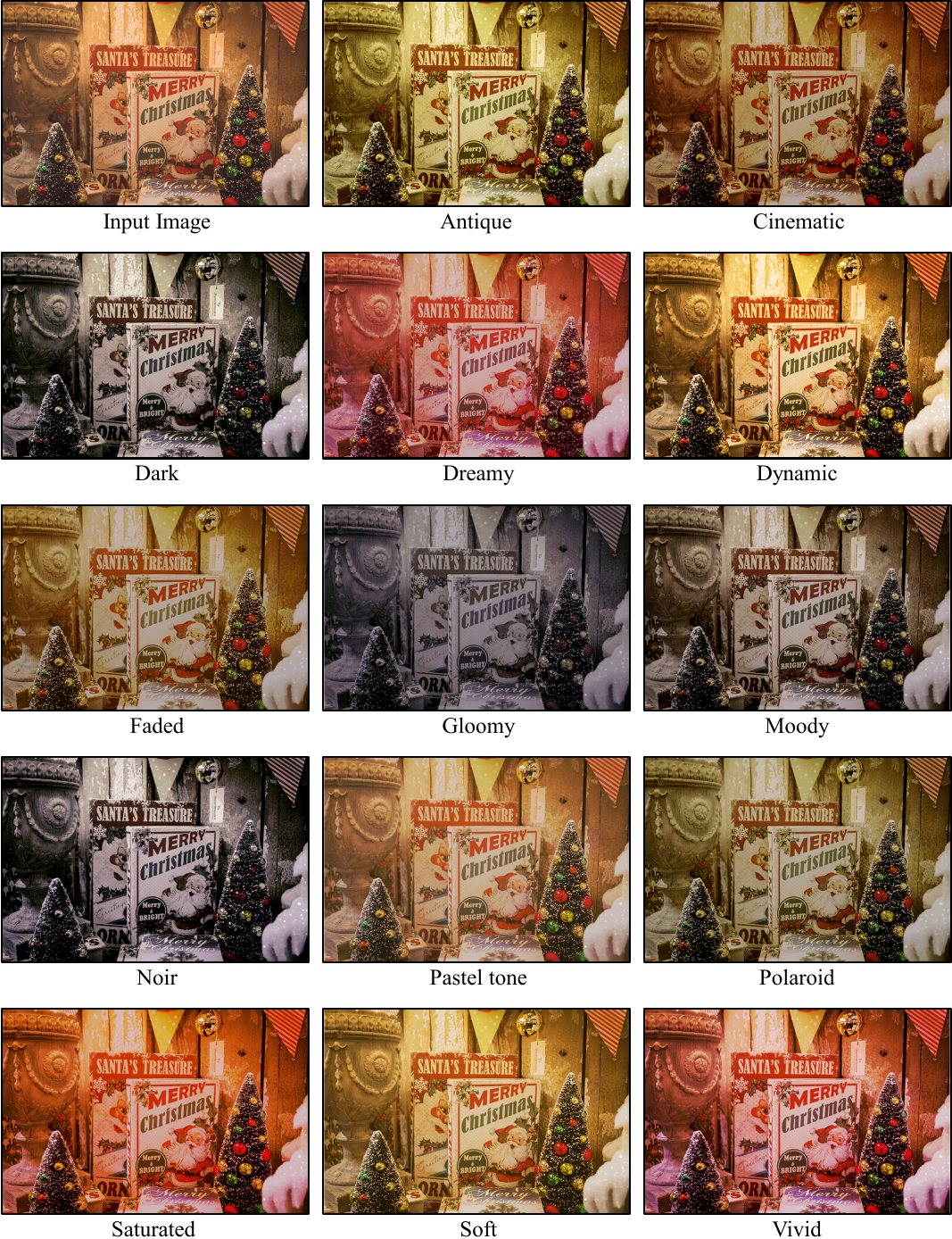}}
    \caption{
    Additional results of \MethodName{} with various text descriptions.
    }
    \label{fig:sup_fig3_examples1}
\end{figure*}

\begin{figure*}[t]
    \centerline{\includegraphics[width=0.95\textwidth]{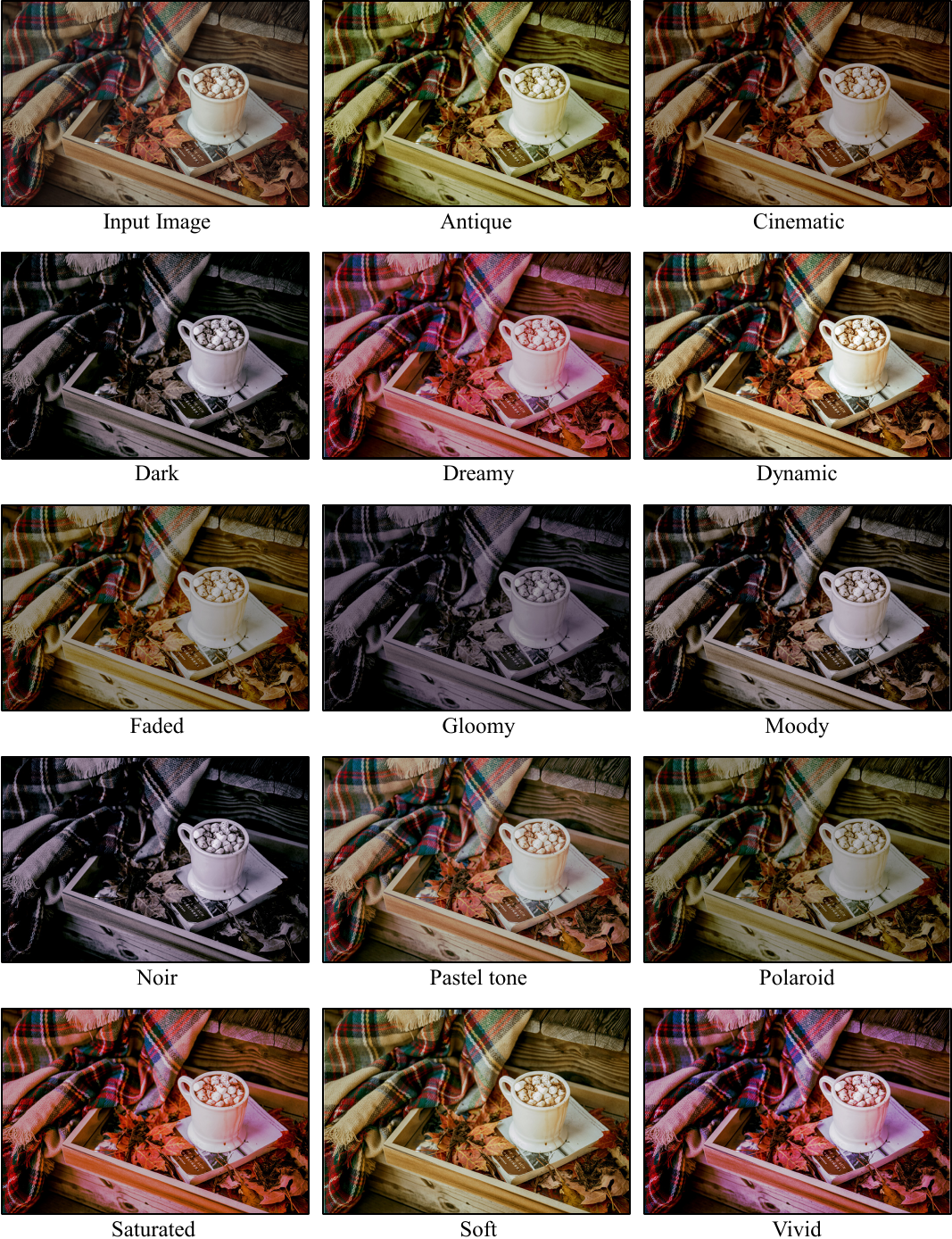}}
    \caption{
    Additional results of \MethodName{} with various text descriptions.
    }
    \label{fig:sup_fig3_examples2}
\end{figure*}

\begin{figure*}[t]
    \centerline{\includegraphics[width=0.95\textwidth]{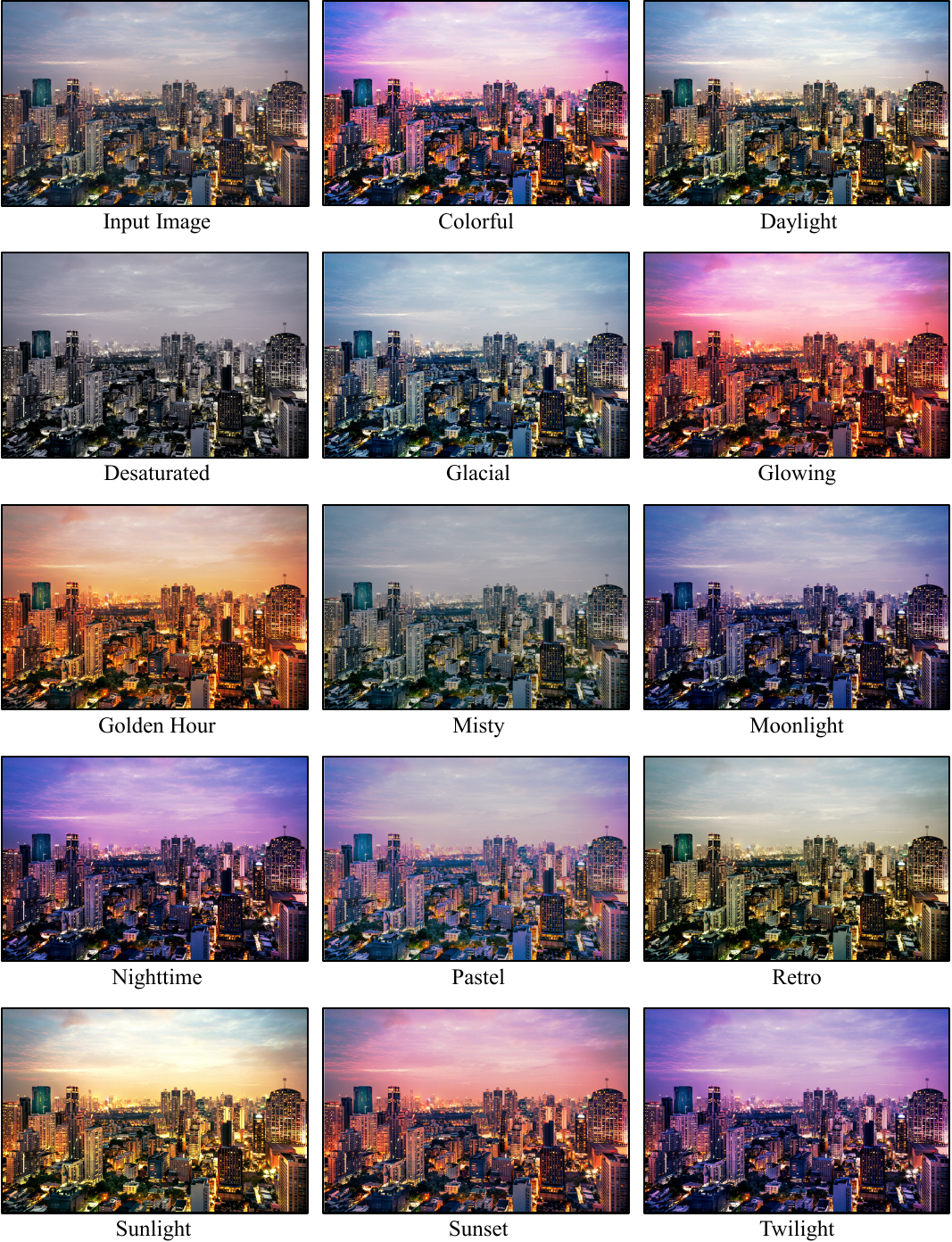}}
    \caption{
    Additional results of \MethodName{} with various text descriptions.
    }
    \label{fig:sup_fig3_examples3}
\end{figure*}

We provide additional results of \MethodName{} with various text descriptions in
\cref{fig:sup_fig3_examples1}, \cref{fig:sup_fig3_examples2} and \cref{fig:sup_fig3_examples3}.
With a comprehensive understanding of natural languages, \MethodName{} supports a wide and diverse range of adjustments, including those previously deemed impossible.

\begin{table*}[t]
\scalebox{0.72}{
\begin{tabular}{c|cc|cc|cc|cc}
\Xhline{3\arrayrulewidth}
CLIP's version &
  \multicolumn{2}{c|}{RN50 ( \MethodName{} )} &
  \multicolumn{2}{c|}{VIT-L/14 ( IP2P )} &
  \multicolumn{2}{c|}{ViT-B/32 ( CLIPstyler )} &
  \multicolumn{2}{c}{RN101 (None) } \\ \hline
Method &
  \multicolumn{1}{c|}{\begin{tabular}[c]{@{}c@{}}CLIP Image\\ Similarity\end{tabular}} &
  \begin{tabular}[c]{@{}c@{}}CLIP Text-Image\\ Direction Similarity\end{tabular} &
  \multicolumn{1}{c|}{\begin{tabular}[c]{@{}c@{}}CLIP Image\\ Similarity\end{tabular}} &
  \begin{tabular}[c]{@{}c@{}}CLIP Text-Image\\ Direction Similarity\end{tabular} &
  \multicolumn{1}{c|}{\begin{tabular}[c]{@{}c@{}}CLIP Image\\ Similarity\end{tabular}} &
  \begin{tabular}[c]{@{}c@{}}CLIP Text-Image\\ Direction Similarity\end{tabular} &
  \multicolumn{1}{c|}{\begin{tabular}[c]{@{}c@{}}CLIP Image\\ Similarity\end{tabular}} &
  \begin{tabular}[c]{@{}c@{}}CLIP Text-Image\\ Direction Similarity\end{tabular} \\ \hline
T2ONet~\cite{t2onet} &
  \multicolumn{1}{c|}{\textbf{0.994}} &
  0.029 &
  \multicolumn{1}{c|}{\textbf{0.994}} &
  -0.001 &
  \multicolumn{1}{c|}{\textbf{0.994}} &
  -0.009 &
  \multicolumn{1}{c|}{\textbf{0.999}} &
  0.030 \\
CLIPstyler~\cite{kwon2022clipstyler} &
  \multicolumn{1}{c|}{0.604} &
  \underline {0.082} &
  \multicolumn{1}{c|}{0.657} &
  0.059 &
  \multicolumn{1}{c|}{0.656} &
  \textbf{0.111} &
  \multicolumn{1}{c|}{0.755} &
  \textbf{0.088} \\
IP2P~\cite{brooks2023instructpix2pix} &
  \multicolumn{1}{c|}{0.915} &
  0.065 &
  \multicolumn{1}{c|}{0.917} &
  \underline {0.060} &
  \multicolumn{1}{c|}{0.928} &
  \underline {0.078} &
  \multicolumn{1}{c|}{0.958} &
  0.074 \\
\MethodName{} &
  \multicolumn{1}{c|}{\underline {0.964}} &
  \textbf{0.089} &
  \multicolumn{1}{c|}{\underline {0.975}} &
  \textbf{0.061} &
  \multicolumn{1}{c|}{\underline {0.966}} &
  0.069 &
  \multicolumn{1}{c|}{\underline {0.988}} &
  \underline {0.084} \\
  \Xhline{3\arrayrulewidth}
\end{tabular}}
\caption{Quantitative comparisons with other versions of CLIP model.
The parentheses alongside each CLIP version denote the methods using the version in their training stage.
}
\label{tab:sup_tbl1_quantitative_comparison}
\end{table*}

{
    \small
    \bibliographystyle{ieeenat_fullname}
    \bibliography{main}
}